\documentclass[lettersize,journal]{IEEEtran}
\usepackage{amsmath,amsfonts}
\usepackage{array}
\usepackage{textcomp}
\usepackage{stfloats}
\usepackage{url}
\usepackage{verbatim}
\usepackage{graphicx}
\usepackage{cite}
\usepackage{url}
\usepackage[hidelinks]{hyperref}
\usepackage{amsmath}
\usepackage{amsthm}
\usepackage{booktabs}
\newcommand{\eg}{{\em e.g.}}		
\newcommand{\ie}{{\em i.e.}}		
\newcommand{\etc}{{\em etc}}		

\usepackage{amsmath,amsfonts,bm}

\def\eqref#1{equation~\ref{#1}}

\def\1{\bm{1}}

\def\rva{{\mathbf{a}}}

\def\rvc{{\mathbf{c}}}
\def\rvd{{\mathbf{d}}}

\def\rvf{{\mathbf{f}}}
\def\rvg{{\mathbf{g}}}
\def\rvh{{\mathbf{h}}}

\def\rvk{{\mathbf{k}}}

\def\rvm{{\mathbf{m}}}

\def\rvo{{\mathbf{o}}}
\def\rvp{{\mathbf{p}}}
\def\rvq{{\mathbf{q}}}

\def\rvs{{\mathbf{s}}}

\def\rvv{{\mathbf{v}}}

\def\rmE{{\mathbf{E}}}

\def\rmR{{\mathbf{R}}}

\def\rmW{{\mathbf{W}}}
\def\rmX{{\mathbf{X}}}

\def\rmZ{{\mathbf{Z}}}

\DeclareMathAlphabet{\mathsfit}{\encodingdefault}{\sfdefault}{m}{sl}
\SetMathAlphabet{\mathsfit}{bold}{\encodingdefault}{\sfdefault}{bx}{n}

\def\gC{{\mathcal{C}}}
\def\gD{{\mathcal{D}}}
\def\gE{{\mathcal{E}}}
\def\gF{{\mathcal{F}}}
\def\gG{{\mathcal{G}}}
\def\gH{{\mathcal{H}}}

\def\gL{{\mathcal{L}}}
\def\gM{{\mathcal{M}}}
\def\gN{{\mathcal{N}}}
\def\gO{{\mathcal{O}}}
\def\gP{{\mathcal{P}}}

\def\gR{{\mathcal{R}}}
\def\gS{{\mathcal{S}}}

\def\gV{{\mathcal{V}}}

\def\gX{{\mathcal{X}}}

\newcommand{\mZero}{{\bf 0}}

\newcommand{\vTheta}{{\bm{\Theta}}}

\newcommand{\withdim}[1]{{\ \in \mathbb{R}^{#1}}}

\usepackage{amsfonts}
\usepackage{bbm}
\usepackage{hyperref}
\usepackage{wasysym}
\usepackage{subcaption}
\usepackage{multirow}
\usepackage{makecell}
\usepackage[ruled,linesnumbered,lined,commentsnumbered]{algorithm2e}
\usepackage{algpseudocode}
\SetKw{KwInput}{Input : }
\SetKw{KwOutput}{Output : }
\SetKw{KwParam}{Parameter : }

\begin{document}

\title{Knowledge-Induced Medicine Prescribing Network for Medication Recommendation}

\author{Ahmad Wisnu Mulyadi and~Heung-Il~Suk,~\IEEEmembership{Senior Member,~IEEE}
\thanks{
A. W. Mulyadi is with the Department of Brain and Cognitive
Engineering, Korea University, Seoul 02841, Republic of Korea (wisnumulyadi@korea.ac.kr).

H.-I. Suk is with the Department of Artificial Intelligence and the
Department of Brain and Cognitive Engineering, Korea University,
Seoul 02841, Republic of Korea, and the corresponding author (e-mail:
hisuk@korea.ac.kr).

}
\thanks{Manuscript received X; revised X.}}

\markboth{Journal of \LaTeX\ Class Files,~Vol.~14, No.~8, August~2021}%
{Shell \MakeLowercase{\textit{et al.}}: A Sample Article Using IEEEtran.cls for IEEE Journals}

\IEEEpubid{0000--0000/00\$00.00~\copyright~2021 IEEE}

\maketitle

\begin{abstract}
Extensive adoption of electronic health records (EHRs) offers opportunities for their use in various downstream clinical analyses. To accomplish this purpose, enriching an EHR cohort with external knowledge (\eg, standardized medical ontology and wealthy semantics) could help us reveal more comprehensive insights via a spectrum of informative relations among medical codes. Nevertheless, harnessing those beneficial interconnections was scarcely exercised, especially in the medication recommendation task. This study proposes a novel Knowledge-Induced Medicine Prescribing Network (KindMed) to recommend medicines by inducing knowledge from myriad medical-related external sources upon the EHR cohort and rendering interconnected medical codes as medical knowledge graphs (KGs). On top of relation-aware graph representation learning to obtain an adequate embedding over such KGs, we leverage hierarchical sequence learning to discover and fuse temporal dynamics of clinical (\ie, diagnosis and procedures) and medicine streams across patients' historical admissions to foster personalized recommendations. Eventually, we employ attentive prescribing that accounts for three essential patient representations, \ie, a summary of joint historical medical records, clinical progression, and the current clinical state of patients. We validated the effectiveness of our KindMed on the augmented real-world EHR cohorts, achieving improved recommendation performances against a handful of graph-driven baselines.
\end{abstract}


\section{Introduction}
\IEEEPARstart{E}{lectronic} health records (EHRs) intensively collect longitudinal medical-associated measurements and codes of patients, making them a valuable resource for various secondary analyses \cite{critical2016secondary,mulyadi2022uncertainty}. Among others, one important use case is leveraging EHRs for personalized medications \cite{bhoi2021premier}. Given assorted patient clinical codes, \ie, diagnosis and procedure codes, this task aims to identify a set of safe and effective medicines that treat the illness tailored to the patient's characteristics and medication history. However, this therapeutic effort remains a non-trivial task due to the risk of drug-drug interactions (DDIs) \cite{tatonetti2012data}, which can decrease the effectiveness of administered medicines in curing the diseases and can raise unexpected adverse effects \cite{qiu2022acomprehensive,smithburger2012ddiinmicu}. Instead of being curative, co-taken drugs with adverse effects could further deteriorate patients' health conditions.

Attempts to account for relations (including DDIs) among medical entities enclosed in EHR have gained significant traction through the employment of medical knowledge graphs (KGs) \cite{chandak2023building,gong2021smr}. Here, graph neural networks (GNNs) have been exploited for medicine recommendations from various perspectives. For instance, GAMENet \cite{shang2019gamenet}, PREMIER \cite{bhoi2021premier}, and COGNet \cite{wu2022cognet} used GNNs to learn embeddings from graphs of drug co-occurrences and DDIs. SafeDrug \cite{yang2021safedrug} employed GNNs to discover both global and local drug features from drugs' atomic bonds and chemical substances, respectively. Similarly, CSEDrug \cite{wu2022CSEDrug} and MoleRec \cite{yang2023molerec} leveraged sub-structure relations of drugs, which are learned by a GNN-based encoding module. Meanwhile, since an EHR comprises patient records across longitudinal admissions, its temporal dynamics can be discovered using sequence models, \eg, recurrent neural networks (RNNs) \cite{choi2016retain,shang2019gamenet} or Transformer \cite{shang2019pretrainingofgraph,sun2022drugrec}, making it a longitudinal model.

Nevertheless, as briefly outlined in Table \ref{table:relation_summary}, those pioneering works mostly overlooked the joint integration of \emph{medical ontology} and \emph{semantic relations} from external knowledge sources. By means of ontology, we can draw a hierarchical structure over medical codes \cite{choi2017gram,falis2019ontological,whetzel2011bioportal}, \eg, international classification of disease (ICD), anatomical therapeutic chemical (ATC) classification. Such \emph{parent-child} relations enable us to discover meaningful coarse-to-fine features over medical concepts \cite{choi2017gram,ma2018kame}. Meanwhile, various semantic relations (\eg, association, interaction, causation \cite{kilicoglu2011constructing}) among medical entities extracted from a massive corpus of biomedical research literature, such as SemMedDB \cite{kilicoglu2012semmedb}, could enhance predictive healthcare and enrich it with reasoning capability \cite{tao2020mining,ye2021medpath}.

\IEEEpubidadjcol
Motivated by this, we engage in the task of medicine recommendation by proposing a novel Knowledge-Induced Medicine Prescribing Network (KindMed), which harnesses various relations applied upon the EHR cohort. Specifically, we extract relations (\ie, ontology, semantics, and DDIs) from medical-related external knowledge to construct admission-wise clinical and medicine KGs for each patient. By utilizing such paired medical KGs, we can model the historical medications of each patient, thereby enabling personalized recommendations. To ensure reliable personalization, we include a node summarizing the patient's demographics (\eg, gender, ethnicity, and age) as an augmentation to the medical KGs. Furthermore, we exploit relation-aware GNNs to learn and aggregate the node embeddings, which are reflected as medical state representations. To capture the influence of a patient's medication history over admission time, we devise a special type of temporal learning called hierarchical sequence learning. This module allows us to learn and fuse the clinical and medicine streams via a collaborative filtering layer by imposing ensemble interaction between streams. Additionally, high-level recurrent networks account for both streams to obtain the joint temporal features. In producing a set of recommended medicine combinations, we employ an attentive prescribing module (APM) built upon an attention mechanism \cite{vaswani2017attention}. This module consumes threefold features from patients' admission records: joint historical medical records summary, clinical progression summary, and the most recent clinical state. To validate the effectiveness of our KindMed, we utilized MIMIC (-III \& -IV) datasets \cite{johnson2016mimic,johnson2023mimic} as the real-world EHR cohorts enriched with extracted relations from external medical knowledge sources \cite{bodenreider2004unified,rindflesch2011semantic,whetzel2011bioportal}. 

In summary, we convey our main contributions as follows:
\begin{itemize}
    \item We propose a novel KindMed model for recommending medicines by harnessing a variety of relationships extracted from external knowledge to construct and model the personalized medical KGs for each patient encounter within the EHR cohort.
    
    \item We account for patients' historical medications by devising a hierarchical sequence learning to independently obtain temporal features over clinical and medicine streams. We then fuse both learned features via a collaborative filtering layer, while relying on a high-level recurrent network to discover the joint temporal features.
    
    \item We introduce APM equipped with an attention mechanism to prescribe medicines by attentively assessing the association between summarized features over joint historical medical records and clinical progression, while also emphasizing the patients' current clinical state.
\end{itemize}

We organize the rest of the article as follows. In Section \ref{section:related_work}, we initiate our discussion by dissecting several closely related works and comparing them to our proposed model in the task of medication recommendation. We outline the problem formulation in Section \ref{section:problem_formulation} before describing our proposed method in Section \ref{section:proposed_method}. In Section \ref{section:experiments}, we thoroughly report our experimental setup and findings followed by discussions for future works stationed under Section \ref{section:discussions}. Ultimately, we conclude this study in Section \ref{section:conclusion}.


\begin{table*}[!t]
\centering
\caption{Brief summary of graph-driven medicine recommender models that harnessed external knowledge.}
\label{table:relation_summary}
\scalebox{1.0}{
\begin{tabular}{@{}lcccccll@{}}
\toprule
\multicolumn{1}{c}{\multirow{2}{*}{\textbf{Model}}} & \multicolumn{3}{c}{\textbf{Relation Type}} & \multicolumn{2}{c}{\textbf{Exerted Upon}} & \multicolumn{1}{c}{\multirow{2}{*}{\textbf{GNN Variant}}} & \multicolumn{1}{c}{\multirow{2}{*}{\textbf{Sequence Model}}} \\ \cmidrule(lr){2-4}\cmidrule(lr){5-6} 
\multicolumn{1}{c}{} & \textbf{Ontology}   & \textbf{Semantics} & \textbf{DDIs} & \textbf{Clinical} & \textbf{Medicine} & \\ \midrule
GAMENet \cite{shang2019gamenet} & \Circle & \rotatebox[origin=c]{90}{\LEFTcircle} & \CIRCLE & \Circle & \CIRCLE & GCN \cite{kipf2017gcn} & GRU \cite{cho2014learning} \\
PREMIER \cite{bhoi2021premier} & \Circle & \rotatebox[origin=c]{90}{\LEFTcircle} & \CIRCLE & \Circle & \CIRCLE & GAT \cite{veličković2018gat} & GRU \cite{cho2014learning} \\
COGNet \cite{wu2022cognet} & \Circle & \rotatebox[origin=c]{90}{\LEFTcircle} & \CIRCLE & \Circle & \CIRCLE & GCN \cite{kipf2017gcn} & Transformer \cite{vaswani2017attention} \\
SafeDrug \cite{yang2021safedrug} & \Circle & \rotatebox[origin=c]{90}{\LEFTcircle} & \CIRCLE & \Circle & \CIRCLE & MPNN \cite{gilmer2017message} & GRU \cite{cho2014learning}\\
CSEDrug \cite{wu2022CSEDrug} & \Circle & \rotatebox[origin=c]{90}{\LEFTcircle}& \CIRCLE & \Circle & \CIRCLE & MPNN \cite{gilmer2017message} & GRU \cite{cho2014learning} \\
G-BERT \cite{shang2019pretrainingofgraph} & \CIRCLE & \Circle & \Circle & \CIRCLE & \CIRCLE & GAT \cite{veličković2018gat} & Transformer \cite{vaswani2017attention}\\
Med-Tree \cite{yue2022medtree} & \CIRCLE & \rotatebox[origin=c]{90}{\LEFTcircle} & \CIRCLE & \CIRCLE & \CIRCLE & GCN, GAT \cite{kipf2017gcn,veličković2018gat} & GRU \cite{cho2014learning}\\
KindMed (Ours) & \CIRCLE & \CIRCLE & \CIRCLE & \CIRCLE & \CIRCLE & R-GCN \cite{schlichtkrull2018modeling} & Hybrid of \cite{cho2014learning,vaswani2017attention}\\
\bottomrule
\multicolumn{8}{l}{\small\Circle: not available; \CIRCLE: fully exercised; \rotatebox[origin=c]{90}{\LEFTcircle}: implicitly exercised, \ie, via co-occurrence of drugs or its chemical sub-structure composition.}
\end{tabular}
}\\
\end{table*}

\section{Related Work}
\label{section:related_work}

\subsection{Graph-Driven Medicine Recommendation} 
Flavors of GNNs \cite{gilmer2017message,kipf2017gcn,veličković2018gat} have been progressively exploited on medical graphs for recommending medicines with diverse construction perspectives. These includes forming interconnections from in-cohort drug co-occurrences \cite{bhoi2021premier,shang2019gamenet,wu2022cognet}, exploring the drugs' chemical sub-structure composition \cite{wu2022CSEDrug,yang2021safedrug}, and accounting for DDIs from external knowledge \cite{bhoi2021premier,gong2021smr,shang2019gamenet,wu2022CSEDrug,wu2022cognet,yang2021safedrug}. To our best knowledge, only a fair few existing works have exploited ontology upon medical entities \cite{shang2019pretrainingofgraph,yue2022medtree} to tackle the medicine recommendation task. In this study, we address the given task with an elevated standpoint in terms of constructing the medical KGs. In particular, we integrate a set of extracted relations under medical ontology (\ie, ICD-9 and ATC), semantics, and DDIs to produce personalized medical KGs, \ie, paired clinical and medicine KG, across patient admissions. Furthermore, unlike existing works that only consider a single relation type between homogeneous medical nodes, our medical KG representation learning takes into account a variety of relations over heterogeneous medical entities to learn and enrich its node embedding via relation-aware GNNs.

\subsection{Longitudinal Approach} 
Meanwhile, closely related works on recommending medications using a longitudinal approach often relied on two main branches of sequence models, \eg, RNNs \cite{bhoi2021premier,shang2019gamenet,wu2022CSEDrug,yang2021safedrug,yue2022medtree} or Transformers \cite{shang2019pretrainingofgraph,sun2022drugrec,wu2022cognet}. Such a model is employed to separately learn the temporal features of diagnoses and procedures over time as main sources of inputs. These features were regarded as patient representations, which were further modeled with graph-driven drug embedding to recommend medicines. By combining diagnosis and procedure codes into a unified clinical KG, we eliminate the need for separate sequence models for each. Instead, we exploit hierarchical sequence learning to discover and jointly fuse the features from the past clinical and medicine streams, yielding personalized medication history modeling. This mechanism allows us access to the learned temporal summaries of joint historical medical records and clinical progression, as well as the recent clinical state of patients. We then dispatch this tuple of key features into our downstream APM, which is built upon a Transformer-based decoding module, to predict a set of recommended medicines for a particular patient.

\section{Problem Formulation}
\label{section:problem_formulation}

\subsection{Electronic Health Records} 
We denote an EHR cohort as $\boldsymbol{\gX} = \{\gX^{n}\}_{n=1}^N$, where a series of records
$\gX^{n} = [\rmX_1^{n}, \ldots, \rmX_t^{n}, \ldots, \rmX^{n}_{T^{n}}]^\top$ is observed for $n$-th patient over their respective total admissions $T^{n}$. Hereafter, we omit the superscript $n$ for clarity. Each admission record $\rmX_t = \{\rvd_t, \rvp_t,\rvm_t,\rvs_t\}$ includes the following components: diagnosis codes $\rvd_t$, procedure codes $\rvp_t$, medicine codes $\rvm_t$, and subject-wise demographics $\rvs_t$. We refer to ${\{\rvd, \rvp, \rvm\}}_t\in \{0,1\}^{\{|\gD|,|\gP|,|\gM|\}}$ as medical multi-hot vectors, with $\gD,\gP,\gM$ denoting their medical entity space and $|\cdot|$ as cardinality.  Using ICD-9 as a designated ontology, we merge diagnosis and procedure codes into clinical codes $\rvc_t \in \{0,1\}^{|\gC|}$, with $\gC = \gD \ \cup \ \gP$. To avoid redundancy, we refer to both medical entities, \ie, clinical and medicine, as $* \in \{\rvc, \rvm\}$. In addition, we utilize demographic vectors $\rvs_t \in \{0,1\}^{K \times |\gS_{1:K}|}$, where $K$ represents auxiliary non-medical data elements (\ie, gender, ethnicity, and age) with its own space $\gS_k$ for $k\in\{1,\ldots,K\}$. We organize a glossary of all utilized notations alongside their concise description in Table \ref{table:notations}.

\begin{table*}[!t]
\centering
\caption{Notations utilized in our proposed KindMed}
\label{table:notations}
\begin{tabular}{@{}cl@{}}
\toprule
\textbf{Notation} & \multicolumn{1}{c}{\textbf{Description}} \\ \midrule
$\boldsymbol{\gX} = \{\gX^{n}\}_{n=1}^N$ & an EHR cohort with $N$ total number of patients \\ 
$\gX^{n} = [\rmX_1^{n}, \ldots, \rmX_t^{n}, \ldots, \rmX^{n}_{T^{n}}]^\top$ & a succession of records for $n$-th patient over their total admissions $T^{n}$ \\
$\mathbf{X}_t=\{\mathbf{d}_t, \mathbf{p}_t, \mathbf{m}_t, \mathbf{s}_t\}$ & $t$-th admission records of a patient (hereafter we dropped the $n$-th patient index) \\ 
$\gD, \gP, \gC=\gD \cup \gP, \gM, \gS$ & entity space of diagnosis, procedure, clinical, medicine and demographics \\
$\mathbf{d}_t \in \{0,1\}^{|\mathcal{D}|}$ & multi-hot vector of diagnosis codes at $t$-th visit\\ 
$\mathbf{p}_t \in \{0,1\}^{|\mathcal{P}|}$ & multi-hot vector of procedure codes at $t$-th visit \\ 
$\mathbf{c}_t \in \{0,1\}^{|\mathcal{C}|}$ & multi-hot vector of unified diagnosis and procedure codes as clinical codes at $t$-th visit\\ 
$\mathbf{m}_t \in \{0,1\}^{|\mathcal{M}|}$ & multi-hot vector of medicine codes at $t$-th visit \\
$\mathbf{s}_t \in \{0,1\}^{K\times|\mathcal{S}_{1:K}|} $ & multi-hot vector of patient demographic information at $t$-th visit \\ 
$*=\{\rvc:=\rvc \cup \rvs, \rvm:=\rvm \cup \rvs \}$ & clinical and medication as two type of medical entities (augmented by demographics) \\
$\gR^{*}$ & relational space of type $*$ \\
$\mathbf{R}_t^{*} \in \mathbb{R}^{|\mathcal{R}^*| \times |*| \times |*|}$ & relational adjacency tensor of type $*$ at $t$-th visit\\ 
$\mathcal{G}_t^{*}=(\mathcal{V}_t^{*}, \mathcal{E}_t^{*}, \mathbf{Z}_t^{*})$ & constructed medical KG of type $*$ at $t$-th visit\\
$\mathcal{V}_t^{*} \in \mathbb{R}^{|*|+1}$ & graph nodes on the medical KG of type $*$ at $t$-th visit\\
$\mathcal{E}_t^{*} \in \mathbb{R}^{|*|+1 \times |\mathcal{R}^*|\times |*|+1}$&  directed edges on the medical KG of type $*$ at $t$-th visit\\
$\mathbf{Z}_t^{*} \in \mathbb{R}^{|*|+1 \times e}$ & node features on medical KG of type $*$ at $t$-th visit\\
$\mathbf{g}_t^{*} \in \mathbb{R}^{e}$ & aggregated node features from medical KG of type $*$ at $t$-th visit \\
$\mathbf{h}_t^{*} \in \mathbb{R}^{e}$ & temporal hidden features of type $*$ at $t$-th visit \\
$\mathbf{f}_t^{\alpha} \in \mathbb{R}^{e}$ & fused features through linear generalized matrix factorization at $t$-th visit \\
$\mathbf{f}_t^{\beta} \in \mathbb{R}^{e}$ & fused features through non-linear deep fusion mechanism at $t$-th visit \\
$\mathbf{f}_t \in \mathbb{R}^{2e}$ & fused features that account for  $\mathbf{f}_t^{\alpha}$ and $\mathbf{f}_t^{\beta}$ at $t$-th visit \\
$\mathbf{h}_t^{\mathbf{c}+\mathbf{m}} \in \mathbb{R}^{e}$ & joint temporal features that accounts both type $*$ at $t$-th visit \\
$\mathbf{o}_t \in \mathbb{R}^{e}$ & output features at $t$-th visit to be fed into layers of predictive FFNs \\
$\hat{\mathbf{m}}_t \in \{0,1\}^{|\mathcal{M}|}$ & multi-hot vector of recommended medicine codes at $t$-th visit \\
\bottomrule
\end{tabular}
\end{table*}

\subsection{Medical KGs Construction} 
For each admission record $\rmX_t$, we construct a pair of disjoint medical KGs \mbox{$\gG^{*}_t = (\gV_t^{*}, \gE_t^{*}, \rmZ_t^*)$}, yielding clinical and medicine KGs. In such graphs, $\gV_t^{*}$, $\gE_t^{*}$, and $\rmZ_t^*$ represent graph nodes, connected edges, and node features, respectively. Here, each observed medical code is considered a node, denoted as $v \in \gV_t^{*}$. To construct the clinical KG $\gG_t^{\rvc}$, we incorporate ancestor nodes for the observed clinical nodes based on the hierarchical structure of ICD ontology, using directed parent-child relations. Additionally, we include semantic relations between these nodes, which are stored in $\rmR_t^{\rvc} \in \mathbb{R}^{|\gR^\rvc|\times|\gC|\times|\gC|}$. Similarly, we consolidate medicine nodes under ATC ontology and DDI relations to construct medicine KG $\gG_t^{\rvm}$, retaining $\rmR^{\rvm}_t \in \mathbb{R}^{|\gR^\rvm|\times|\gM|\times|\gM|}$. These relational adjacency tensors $\rmR_t^{*}$ preserve the indices of interlinked clinical and medicine codes, with the respective relation space $\gR^{*}$ extracted from medical-related external knowledge. For initializing the node features, we transform each utilized medical and demographic multi-hot vector into their respective embeddings, expressed as follows
\begin{align}
    \rmZ_t^\rvc &= \rmE_{\rvc}\rvc_t, \\
    \rmZ_t^\rvm &= \rmE_{\rvm}\rvm_t, \\
    \rmZ_t^\rvs &= \textstyle\sum_{k=1}^K \rmE_{\rvs_k}\rvs_t[k],
\end{align}
\noindent where $\rmE_{\{\rvc,\rvm, \rvs_{1:K}\}}\withdim{e \times \{|\gC|,|\gM|, |\gS_{1:K}|\}}$ denotes the embedding matrices, with $e$ indicating the dimension of embeddings. Since demographic features contain multiple pieces of information, we aggregate their embedded features to form a single subject-wise demographic feature vector. We assign each node in $\gV_t^{*} \withdim{|*|+1}$ with these embedded features $\rmZ_{t}^{*}[v] \withdim{e}$, $\forall \ v \in \gV_t^{*}$. Augmenting the set of nodes by one is necessary due to the inclusion of demographic features as a master node \cite{gilmer2017message}, such that $\rvc:=\rvc \cup \rvs$ and $\rvm:=\rvm \cup \rvs$. This also implies adding auxiliary relations $\rmR_t^{\rvs} \withdim{1 \times |\gV_t^{*}| \times |\gV_t^{*}|}$ from the originally observed medical codes in the EHR cohort 
 (\ie, leaf nodes in $\gG^{*}_t$) to this patient node. Lastly, to represent the relationships between nodes in our medical knowledge graphs, we define a directed edge from a node $u$ to a node $v$ using relation $r^{*}\in \gR^{*}$, for all node pairs $\forall \ (u,v) \in \gV_t^{*}$, formulated as follows
\begin{align}
    \gE^{*}_t[u,r^{*},v] &= \begin{cases}
  1, & \text{if relation exists in } \rmR_t^{*}[r^{*},u,v], \\
  0, & \text{otherwise,}
\end{cases}.
\end{align}

\subsection{Objective} 
The main goal of this study is to recommend medications for patients within a longitudinal EHR cohort by utilizing medical KGs spanning their admission records. Specifically, considering the longitudinal healthcare records of patients represented by the most recent clinical KG $\gG_t^{\rvc}$ and their previous medical KGs $\gG_{1:t-1}^{*}$, our proposed KindMed aims to predict a set of prescribed medicines $\hat{\rvm}_t$ by modeling $p(\hat{\rvm}_t|\gG_t^{\rvc}; \gG_{1:t-1}^{*})$.

\begin{figure*}
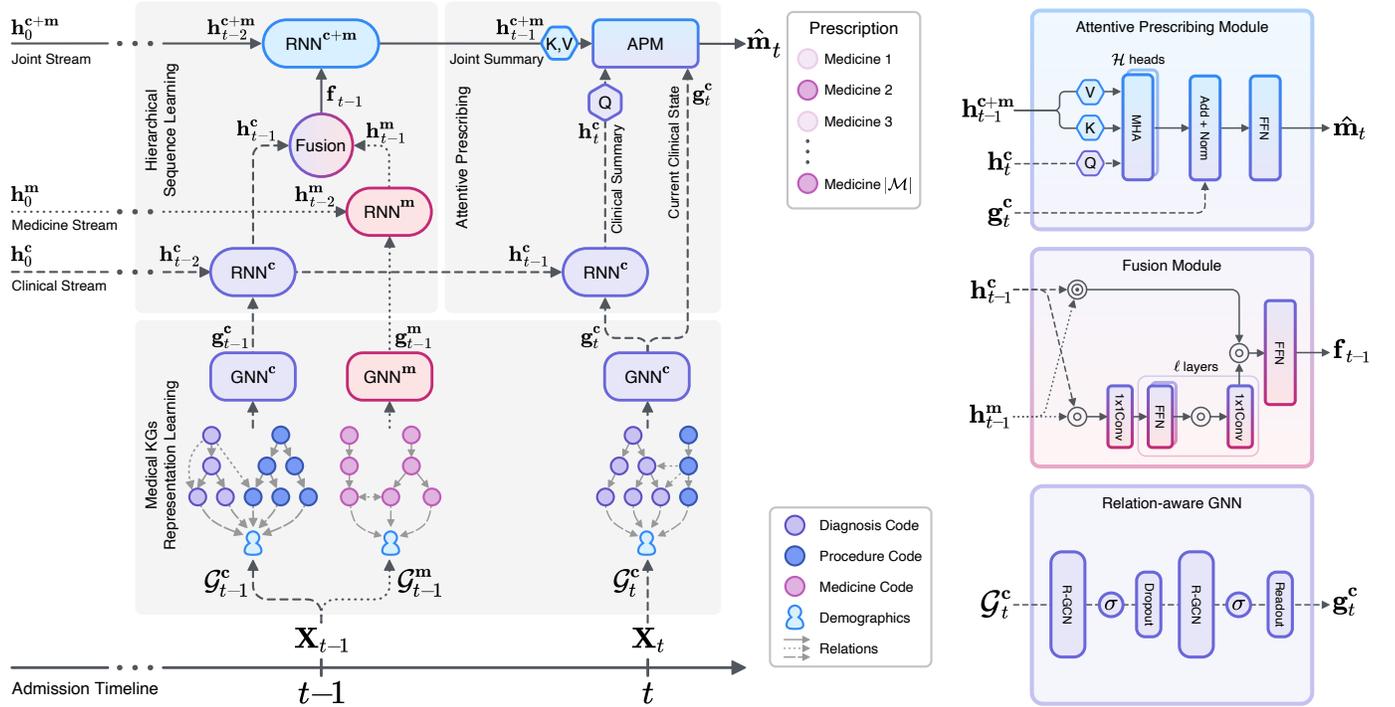

\centering
\begin{subfigure}{0.68\textwidth}
    \begin{minipage}[b][][t]{1.0\linewidth}
        \includegraphics[width=1.0\textwidth]{figures/kindmed.pdf}
    \end{minipage}
\end{subfigure}
\hfill
\begin{subfigure}{0.3\textwidth}
    \begin{minipage}[b][][t]{1.0\linewidth}
        \includegraphics[width=1.0\textwidth]{figures/kindmed_modules.pdf}
    \end{minipage}
\end{subfigure}
\caption{Overview of KindMed and its internal modules, including relation-aware GNNs for learning and enriching the node embedding in medical KGs, a fusion module for integrating temporal features from clinical and medicine streams, and an attentive prescribing module for recommending medications.}
\label{fig:kindmed_model}
\end{figure*}

\section{Proposed Method of KindMed}
\label{section:proposed_method}

We illustrate our proposed KindMed model in Figure \ref{fig:kindmed_model}, consisting of three primary components: (i) \emph{medical KGs representation learning} for obtaining medical state representations; (ii) \emph{hierarchical sequence learning} for discovering joint temporal features from the clinical and medicine streams; and eventually, (iii) \emph{attentive prescribing} for producing a set of recommended medicines.

\subsection{Medical KGs Representation Learning}
Our constructed medical KGs consist of multiple medical entities, including diagnoses, procedures, medicine nodes, as well as auxiliary patient nodes in $\gG^{*}_t$, interconnected by a wide variety of relations in $\gE^{*}_t$. Therefore, in structurally learning these medical KG representations, we need to utilize variants of GNN that consider these aspects. Among several options meeting this criterion, we select the relational graph convolutional network (R-GCN) \cite{schlichtkrull2018modeling}, which is specifically designed to model multi-relational data and learn the node embeddings. At the $l$-th layer of R-GCN, we update the node features $\rmZ_{t}^{*,l}[v]$ by taking into account their interconnected relations, which are formulated as
\begin{align}
    \label{eq:gnn_rgcn}
    \rmZ_{t}^{*,l}[v] = \sigma  \Biggl( \sum_{r^*}^{\gR^*} \sum_{u}^{\gN^{r^{*}}_v}  \frac{1}{\eta}\rmW_{r^*}^{l'} \rmZ_{t}^{*,l'}[u] + \rmW_{*}^{l'} \rmZ_{t}^{*,l'}[v] \Biggr),
\end{align}
where $l'=l-1$, $\gN^{r^{*}}_v=\{u \in \gV^*_t \mid (u,r^*,v) \in\gE_t^*\}$ represents a set of nodes one-hop away toward node $v$ under relation $r^*$, with $\eta=|\gN^{r^{*}}_v|$ denoting the number of nodes in this set. The update involves learnable projection matrices $\rmW_{r^*}^{l'}, \rmW_{*}^{l'} \withdim{e \times e}$, where $e$ denotes the embedding size. Additionally, $\sigma$ indicates a non-linear activation function. We omit the respective biases for clarity throughout the article. This process entails a nonlinear transformation of each node embedding and aggregation over a set of source nodes toward their target nodes (including themselves) under a specific relation. Finally, the aggregated representations $\rvg_t^{*} \withdim{e}$ for each medical KG can be induced through a readout equation, as expressed in the following Eq. (\ref{eq:gnn_readout}).
\begin{align}
    \label{eq:gnn_readout}
    \rvg_t^{*} = \sum_{v}^{\gV_t^{*}} \rmZ_{t}^{*,l}[v].
\end{align}
By processing the medical KGs across admissions in this manner, we derive a tuple of learned medical state representations $\{\rvg_{1:t}^{\rvc}, \rvg_{1:t-1}^{\rvm}\}$, which are subsequently leveraged in the hierarchical sequence learning.

\subsection{Hierarchical Sequence Learning}
Using the representations $\{\rvg_{1:t}^{\rvc}, \rvg_{1:t-1}^{\rvm}\}$ obtained by relation-aware GNNs, we introduce hierarchical RNNs to (i) independently learn the temporal dynamics of clinical and medicine streams via respective $\textrm{RNN}^{\rvc}$ and $\textrm{RNN}^{\rvm}$; (ii) fuse both temporal features via a collaborative filtering layer; and ultimately (iii) discover joint temporal hidden features of the fused representations via $\textrm{RNN}^{\rvc + \rvm}$ up to the penultimate admission. 

Foremost, we learn the temporal hidden features $\rvh^{*}\withdim{e}$ of medical records across admissions by employing corresponding $\textrm{RNN}^{*}$ for the clinical and medicine streams through
\begin{align}
    \rvh^{\rvc}_{t} &= \textrm{RNN}^{\rvc}(\rvg^{\rvc}_{1:t}; \rvh^{\rvc}_{1:t-1}),\\
    \rvh^{\rvm}_{t-1} &= \textrm{RNN}^{\rvm}(\rvg^{\rvm}_{1:t-1}; \rvh^{\rvm}_{1:t-2}).
\end{align}
We use the term RNN to refer to any sequence model capable of effectively processing sequential data. Specifically, we employ the gated-recurrent unit (GRU) cell \cite{cho2014learning} for learning the hidden features sequentially. For each admission prior to $t$-th, we utilize a fusion module with a modified neural collaborative filtering layer \cite{he2017neuralcf} to combine both streams of temporal features $\{\rvh^{\rvc}_{1:t-1},\rvh^{\rvm}_{1:t-1}\}$ through ensemble interaction functions expressed as 
\begin{align}
    \label{eq:fusion_gmf}
    \rvf^{\alpha}_{t-1} &= \rvh^{\rvc}_{t-1} \odot \rvh^{\rvm}_{t-1}, \\ 
    \label{eq:fusion_ffn}
    \rvf^{\beta}_{t-1} &= \gF^{1:\ell}([\rvh^{\rvc}_{t-1} \circ \rvh^{\rvm}_{t-1}]),\\
        \label{eq:fusion_merge}
    \rvf_{t-1} &= \rmW_{\rvf}^2 (\sigma(\rmW_{\rvf}^1[ \rvf^{\alpha}_{t-1} \circ \rvf^{\beta}_{t-1} ])).
\end{align}
Here, we denote $\odot$ as the Hadamard product and $\circ$ as the concatenation operator. We directly multiply those features to simulate the generalized matrix factorization in obtaining $\rvf^{\alpha}_{t-1}$. Simultaneously, we acquire the features $\rvf^{\beta}_{t-1}$ through $\gF^{1:\ell}$, indicating $\ell$ layers of feed-forward networks (FFNs) equipped with a deep fusion mechanism \cite{chen2017multi}. This mechanism involves a series of $1\times1$ convolutions followed by a non-linearity in the intermediate layers to impose more interactions between input features. We further combine the concatenated features of $\rvf^{\alpha}_{t-1}$ and $\rvf^{\beta}_{t-1}$  with projection matrices $\rmW_{\rvf}^{\{1,2\}} \withdim{2e \times 2e}$ to incorporate both linear and non-linear features. Each admission-wise fused feature set $\rvf_{1:t-1}$ will be fed into the high-level recurrent networks $\textrm{RNN}^{\rvc+\rvm}$ to learn joint temporal hidden features $\rvh^{\rvc+\rvm}_{1:t-1}\withdim{e}$ as
\begin{align}
    \rvh^{\rvc+\rvm}_{t-1} = \textrm{RNN}^{\rvc+\rvm}(\rvf_{1:t-1}; \rvh^{\rvc+\rvm}_{1:t-2}).
\end{align}


\subsection{Attentive Prescribing} 
To this end, we obtain a tuple of patient features $\{\rvh^{\rvc}_t, \rvh^{\rvc+\rvm}_{t-1}\}$ that capture the temporal dynamics of clinical progressions and joint historical medication records. These features serve as the foundation for making accurate and personalized recommendations based on the patient's medication history. Additionally, we argue that the graph-level aggregated features $\rvg^{\rvc}_{t}$, representing the current clinical state, could also be crucial for recommending medicines, independent from those temporal features. For instance, in cases where a patient's illness at the current admission is unprecedented, neither clinical nor medication records from past admissions would be relevant. To address such scenarios, we exploit an attention mechanism through a Transformer-based decoder \cite{vaswani2017attention}, which calibrates the patient features to optimize the downstream recommendation task. 

Given a set of features designated as query $\rvq$, key $\rvk$, and value $\rvv$, we derive the attended features using scaled-dot product attention $\phi(\rvq, \rvk, \rvv)$ in Eq. (\ref{eq:mha_sdp}). This process can be extended to enhance expressiveness with a predefined number $\gH$ of projecting heads, resulting in multi-head attention (MHA) $\Phi(\rvq, \rvk, \rvv; \gH)$ as expressed in Eq. (\ref{eq:mha_mha1})-(\ref{eq:mha_mha2}). 
\begin{align}
    \label{eq:mha_sdp}
    \phi(\rvq, \rvk, \rvv) &= \textrm{Softmax}\left( \frac{\rvq\rvk^{\top}}{\sqrt{d_k}}\right) \rvv, \\
    \label{eq:mha_mha1}
    \rva^{h} &= \phi(\rmW_{\rvq}^h\rvq, \rmW_{\rvk}^h\rvk, \rmW_{\rvv}^h\rvv), \\
    \label{eq:mha_mha2}
    \Phi(\rvq, \rvk, \rvv; \gH) &= \rmW_{\Phi}[\rva^1 \circ \ldots \rva^h \ldots \circ \rva^{\gH}].
\end{align}
We adopt this attention mechanism into APM by incorporating our learned features as $\Phi(\rvh^{\rvc}_t, \rvh^{\rvc+\rvm}_{t-1}, \rvh^{\rvc+\rvm}_{t-1}; \gH)$ to simulate the diagnostic process of a clinician, wherein they attentively evaluate the progression of clinical conditions in comparison to the patient's historical medication records. It is worth noting that an alternative approach could involve utilizing a sequence of joint temporal features $\rvh^{\rvc+\rvm}_{1:t-1}$ as keys and values, allowing the entire history of past admission records to be attended. Additionally, to incorporate and underscore the current clinical state, we introduce an additive residual connection with respect to $\rvg^{\rvc}_{t}$, followed by a layer normalization (LNorm), as shown in Eq. (\ref{eq:apm_residual_norm}). Subsequently, the resulting features $\rvo_t$ are passed through a predictive FFN equipped with a set of weight matrices $\rmW_\rvo^1\withdim{2e\times e}, \rmW_\rvo^2\withdim{|\gM|\times 2e}$, and concluded with a sigmoid activation $\bar{\sigma}$ to recommend a set of medicines $\hat{\rvm}_t$ as follows
\begin{align}
    \label{eq:apm_residual_norm}
    \rvo_t &= \textrm{LNorm}(\rvg^{\rvc}_t + \Phi (\rvh^{\rvc}_t,\ \rvh^{\rvc+\rvm}_{t-1}, \ \rvh^{\rvc+\rvm}_{t-1}; \gH))\\
    \label{eq:apm_prescribe}
    \hat{\rvm}_t &= \bar{\sigma}(\rmW_\rvo^2\sigma(\rmW_\rvo^1(\rvo_t)))
\end{align}


\subsection{Objective Functions} 
By regarding this medication recommendation task as multi-label prediction, we optimize our proposed KindMed model via a binary cross-entropy loss $\mathcal{L}_{\textrm{bce}}$ and multi-label margin loss $\mathcal{L}_{\textrm{multi}}$ formulated as
\begin{align}
    \mathcal{L}_{\textrm{bce}} &= - \sum_{t=1}^T \sum_{i=1}^{|\gM|} \rvm_t[i]\log\hat{\rvm}_t[i] + (1-\rvm_t[i])\log(1-\hat{\rvm}_t[i]), \\ 
    \mathcal{L}_{\textrm{multi}} &= \sum_{t=1}^T  \sum_{i,j:\rvm_t[i]=1,\rvm_t[j]=0} \frac{\textrm{max}\left(0, 1 - (\hat{\rvm}_t[i] -  \hat{\rvm}_t[j])\right)}{|\gM|}.
\end{align}
In addition, in the field of medicine recommendation research, an auxiliary DDI loss has been utilized alongside the primary prediction loss to constrain the DDI rate, serving as a measure of medication safety (\emph{i.e.,} aiming for a lower DDI rate). This approach has been previously reported in pioneering models such as \cite{shang2019gamenet,yang2021safedrug}. The external information used for this purpose is extracted from the TWOSIDES database \cite{tatonetti2012data}, which stores pairs of drugs known to cause side effects due to their interactions. Therefore, we incorporate a DDI loss $\mathcal{L}_{\textrm{DDI}}$ to ensure medication safety when recommending a set of medicines $\hat{\rvm}_t$, referencing a DDI adjacency matrix $\rmR_{\textrm{DDI}} \withdim{|\gM|\times|\gM|}$ obtained from those external knowledge, estimated as
\begin{align}
    \mathcal{L}_{\textrm{DDI}}  &= \sum_{t=1}^T  \sum_{i=1}^{|\gM|} \sum_{j=1}^{|\gM|} \rmR_{\textrm{DDI}}[i,j]  \ . \ \hat{\rvm}_{t}[i] \ . \ \hat{\rvm}_{t}[j].
\end{align}
Finally, we formulated the overall loss $\mathcal{L}_{\textrm{total}}$ to optimize all learnable parameters across all modules of the KindMed model in an end-to-end manner. This loss function is calculated as 
\begin{align}
    \label{eq:loss_total}
    \mathcal{L}_{\textrm{total}} &= \lambda_{\textrm{DDI}}(\lambda_{\textrm{rec}} \mathcal{L}_{\textrm{bce}} + (1-\lambda_{\textrm{rec}})\mathcal{L}_{\textrm{multi}}) + (1 - \lambda_{\textrm{DDI}})\mathcal{L}_{\textrm{DDI}},
\end{align}
where  $\lambda_{\textrm{rec}}$ denotes a hyperparameter to weigh the recommendation losses. Following the approach of \cite{yang2021safedrug} to adeptly regulate the DDI loss, we further estimate a weight parameter $\lambda_{\textrm{DDI}}$ using via Eq. (\ref{eq:lambda_ddi}) based on the observed patient-level DDI rate $\delta_{\textrm{DDI}}$, with a controllable DDI threshold $\tau$ and hyperparameter $\rho$. 
\begin{align}
    \label{eq:lambda_ddi}
    \lambda_{\textrm{DDI}} = \begin{cases} 1, &\quad \delta_{\textrm{DDI}}\leq\tau \\\textrm{max}\{0, 1-\frac{\normalsize{\delta_{\textrm{DDI}}}- \normalsize{\tau}}{\normalsize{\rho}}\},&\quad \textrm{otherwise.}\end{cases}
\end{align}

We present the complete training algorithm for our proposed KindMed model in Algorithm \ref{algo:kindmed}.

\begin{algorithm}[!h]
    \SetAlgoLined
    \SetKwInOut{Input}{input}\SetKwInOut{Output}{output}
    \KwInput{\emph{Patient records} $\rmX_{1:T}, \rmR^{*}_{1:T}$} \\
    \KwParam{\emph{Model parameter} $\vTheta$\emph{,} $\{\gH, \tau, \rho, \lambda_{\emph{\textrm{rec}}}\}$}\\
    \KwOutput{\emph{Recommended medicines} $\hat{\rvm}_{1:T}$}\\
    \While{not converge}{
        \For{$t\leftarrow 1$ \KwTo $T$ }{
            $\{\rvh^{*}_0, \rvh^{\rvc+\rvm}_0\} \leftarrow \mZero$ \\
            \For{$\iota\leftarrow 1$ \KwTo $t$ }{
                $\rmZ^{*}_\iota \leftarrow $ \emph{Apply embedding using \emph{Eq. (1)-(3)}}  \\
                $\gG^{*}_\iota \leftarrow $ \emph{Make KGs using} $\rmX^{*}_\iota$, $\rmZ^{*}_\iota$ \emph{and} $\rmR^{*}_\iota$ \\
                $\rmZ_{\iota}^{\rvc,L} \leftarrow \textrm{R-GCN}^{1:L}(\gG_{\iota}^{\rvc})$ \\
                $\rvg^{\rvc}_\iota \leftarrow \textstyle\sum_{v \in \gV_\iota^{\rvc}} \rmZ_{\iota}^{\rvc,L}[v]$ \\
                $\rvh^{\rvc}_\iota \leftarrow \textrm{RNN}^{\rvc}(\rvg^{\rvc}_\iota;\rvh^{\rvc}_{\iota-1})$ \\
                \If{$\iota \leq t-1$}{
                            $\rmZ_{\iota}^{\rvm,L} \leftarrow \textrm{R-GCN}^{1:L}(\gG_{\iota}^{\rvm})$ \\
                $\rvg^{\rvm}_\iota \leftarrow \textstyle\sum_{v\in \gV_\iota^{\rvm}} \rmZ_{\iota}^{\rvm,L}[v]$ \\
                $\rvh^{\rvm}_\iota \leftarrow \textrm{RNN}^{\rvm}(\rvg^{\rvm}_\iota;\rvh^{\rvm}_{\iota-1})$ \\   $\rvf_\iota \leftarrow $ \emph{Employ fusion in \emph{Eq. (9-11)}}\\
                $\rvh^{\rvc+\rvm}_\iota \leftarrow \textrm{RNN}^{\rvc+\rvm}(\rvf_\iota;\rvh^{\rvc+\rvm}_{\iota-1})$ 
                }
            }
            $\rvo_t \leftarrow \textrm{LNorm}(\rvg^{\rvc}_t + \Phi (\rvh^{\rvc}_t,\ \rvh^{\rvc+\rvm}_{t-1}, \ \rvh^{\rvc+\rvm}_{t-1}; \gH))$\\
            $\hat{\rvm}_t \leftarrow \bar{\sigma}(\rmW_\gO^2\sigma(\rmW_\gO^1(\rvo_t)))$ \\
            $\gL_{\textrm{total}} \leftarrow $ \emph{Calculate total loss using} Eq. (21) \\
            $\nabla_{\vTheta} \leftarrow \frac{\partial}{\partial{\vTheta}} \gL_{\textrm{total}}$ \\
            ${\vTheta} \leftarrow \textrm{optimize}({\vTheta},\nabla_{\vTheta})$ \\
        }
    }
    \caption{KindMed for a patient records $\gX^{n}$}
    \label{algo:kindmed}
\end{algorithm}



\section{Experiments}
\label{section:experiments}

\subsection{Dataset} 
Following the preprocessing steps outlined in \cite{shang2019gamenet,yang2021safedrug} for the MIMIC-III \cite{johnson2016mimic} database, we applied several criteria for cohort selection. This included filtering the most commonly occurring clinical codes and screening the medications prescribed within the first 24 hours of admission. In constructing medical KGs, we harnessed several sources of medical-related external knowledge \cite{bodenreider2004unified,kilicoglu2012semmedb,tatonetti2012data,whetzel2011bioportal}. Specifically, for each observed diagnosis and procedure code, we included its ancestor codes based on the ICD-9 Clinical Modification ontology\footnote{\href{https://bioportal.bioontology.org/ontologies/ICD9CM}{https://bioportal.bioontology.org/ontologies/ICD9CM}}. We incorporated semantic relations between these clinical codes using SemMedDB \cite{kilicoglu2012semmedb}\footnote{\href{https://lhncbc.nlm.nih.gov/ii/tools/SemRep_SemMedDB_SKR.html}{https://lhncbc.nlm.nih.gov/ii/tools/SemRep\_SemMedDB\_SKR.html}}, which utilizes the concept unique identifier (CUI) from the unified medical language system (UMLS) \cite{bodenreider2004unified}. By doing this mapping process, we could map ICD-9 codes to CUI and eventually get the semantic relations in SemMedDB. As for the prescriptions, which were originally recorded using the National Drug Code (NDC)), each selected medicine code was first converted into the third level of ATC\footnote{\href{https://bioportal.bioontology.org/ontologies/ATC}{https://bioportal.bioontology.org/ontologies/ATC}}. We then incorporated its ontology relations up to the first level. Note that we included ancestor nodes for modeling the past medications and considered the leaf nodes as the ground truth of the medications at the last admission. Finally, we extracted DDIs to account for the side effects between those medicine codes according to the TWOSIDES \cite{tatonetti2012data}, focusing on the top 40 severity types.

\begin{figure}[!t]
\centering
\begin{subfigure}{0.32\linewidth}
    \begin{minipage}[b][][t]{1.0\linewidth}
        \includegraphics[width=1.0\textwidth]{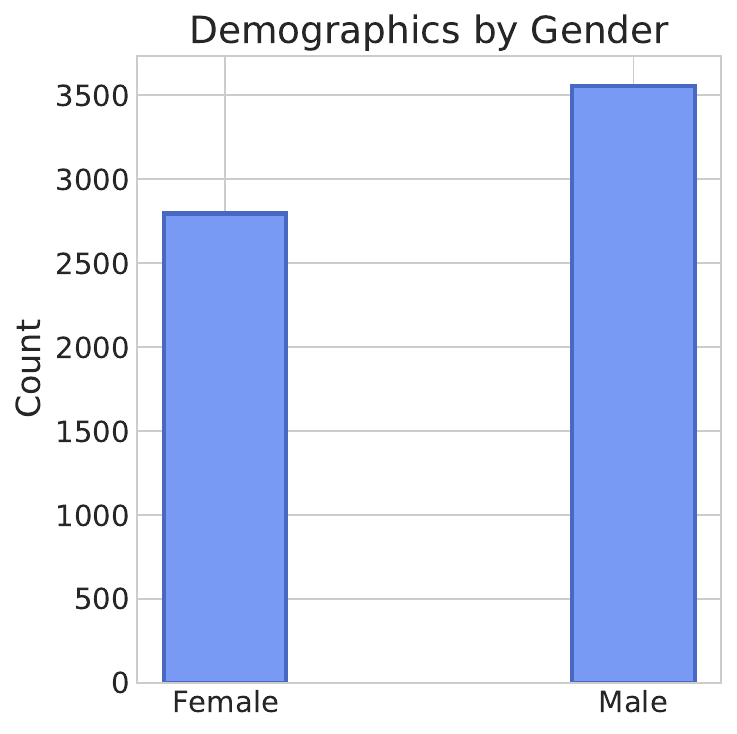}
        \subcaption{Gender}
        \label{fig:demographics_gender}
    \end{minipage}
\end{subfigure}
\hfill
\begin{subfigure}{0.66\linewidth}
    \begin{minipage}[b][][t]{1.0\linewidth}
        \includegraphics[width=1.0\textwidth]{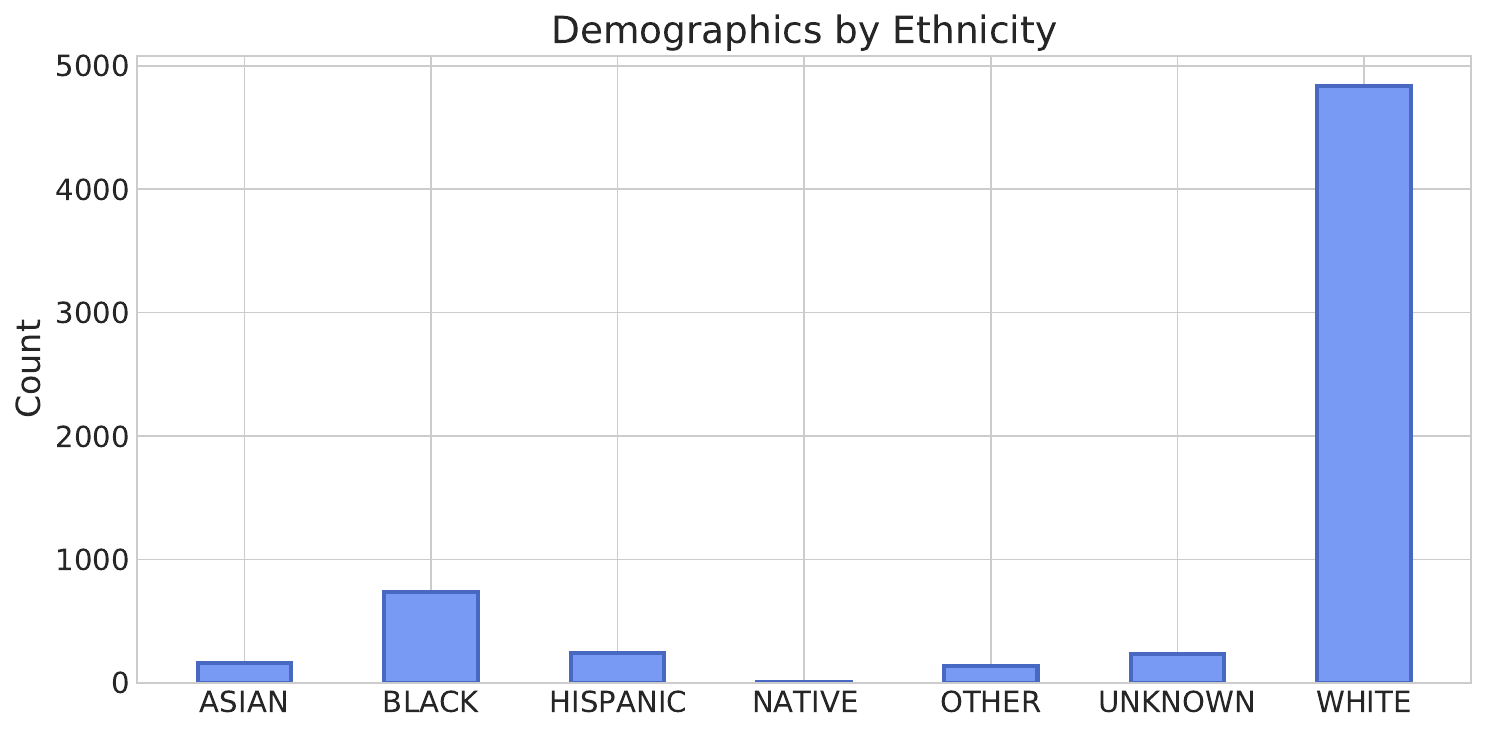}
        \subcaption{Ethnicity}
        \label{fig:demographics_ethnicity}
    \end{minipage}
\end{subfigure}
\hfill
\begin{subfigure}{1.0\linewidth}
    \begin{minipage}[c][][t]{1.0\linewidth}
        \includegraphics[width=1.0\textwidth]{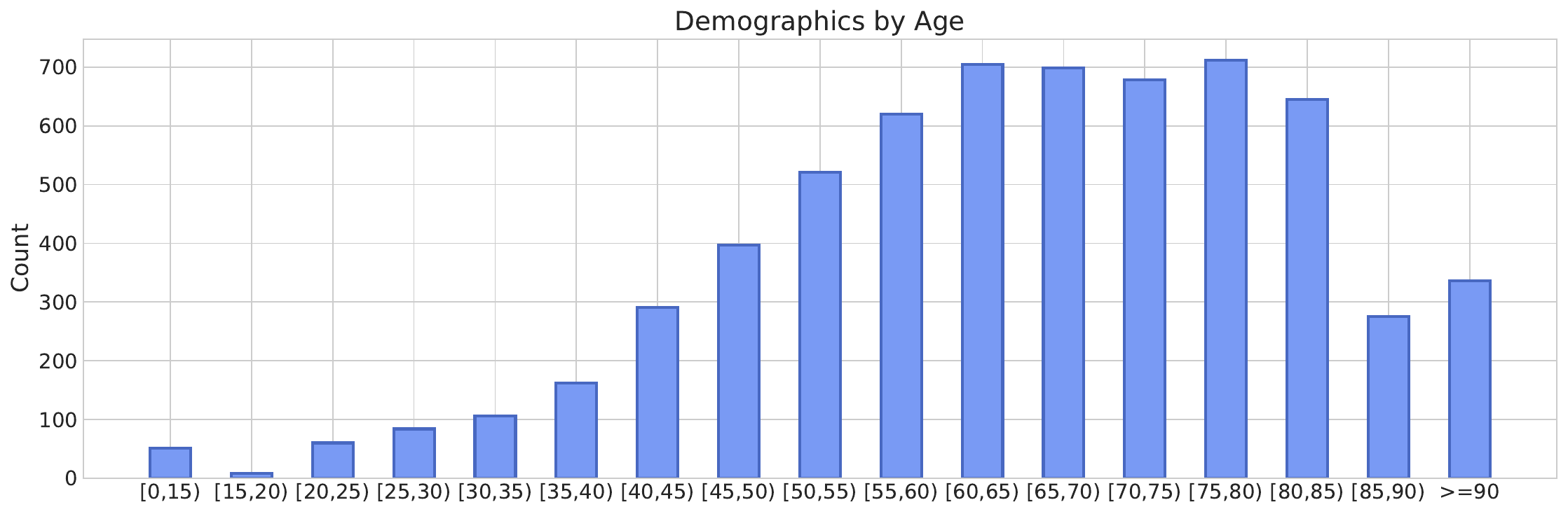}
        \subcaption{Age}
        \label{fig:demographics_age}
    \end{minipage}
\end{subfigure}
\caption{Summary of demographic bins in MIMIC-III cohort.}
\label{fig:demographics_all}
\end{figure}

\begin{figure}[!t]
\centering
\begin{subfigure}{0.32\linewidth}
    \begin{minipage}[b][][t]{1.0\linewidth}
        \includegraphics[width=1.0\textwidth]{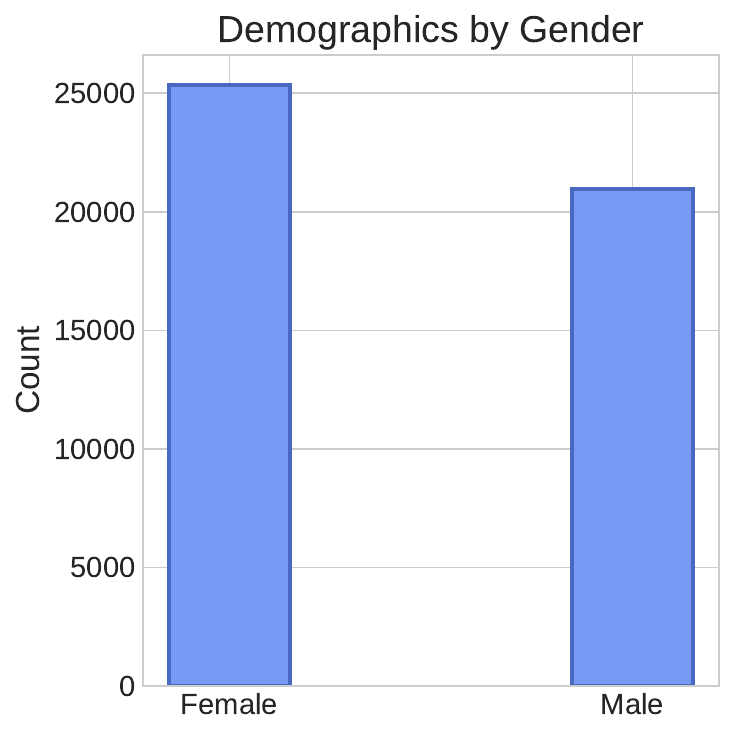}
        \subcaption{Gender}
        \label{fig:mimiciv_demographics_gender}
    \end{minipage}
\end{subfigure}
\hfill
\begin{subfigure}{0.66\linewidth}
    \begin{minipage}[b][][t]{1.0\linewidth}
        \includegraphics[width=1.0\textwidth]{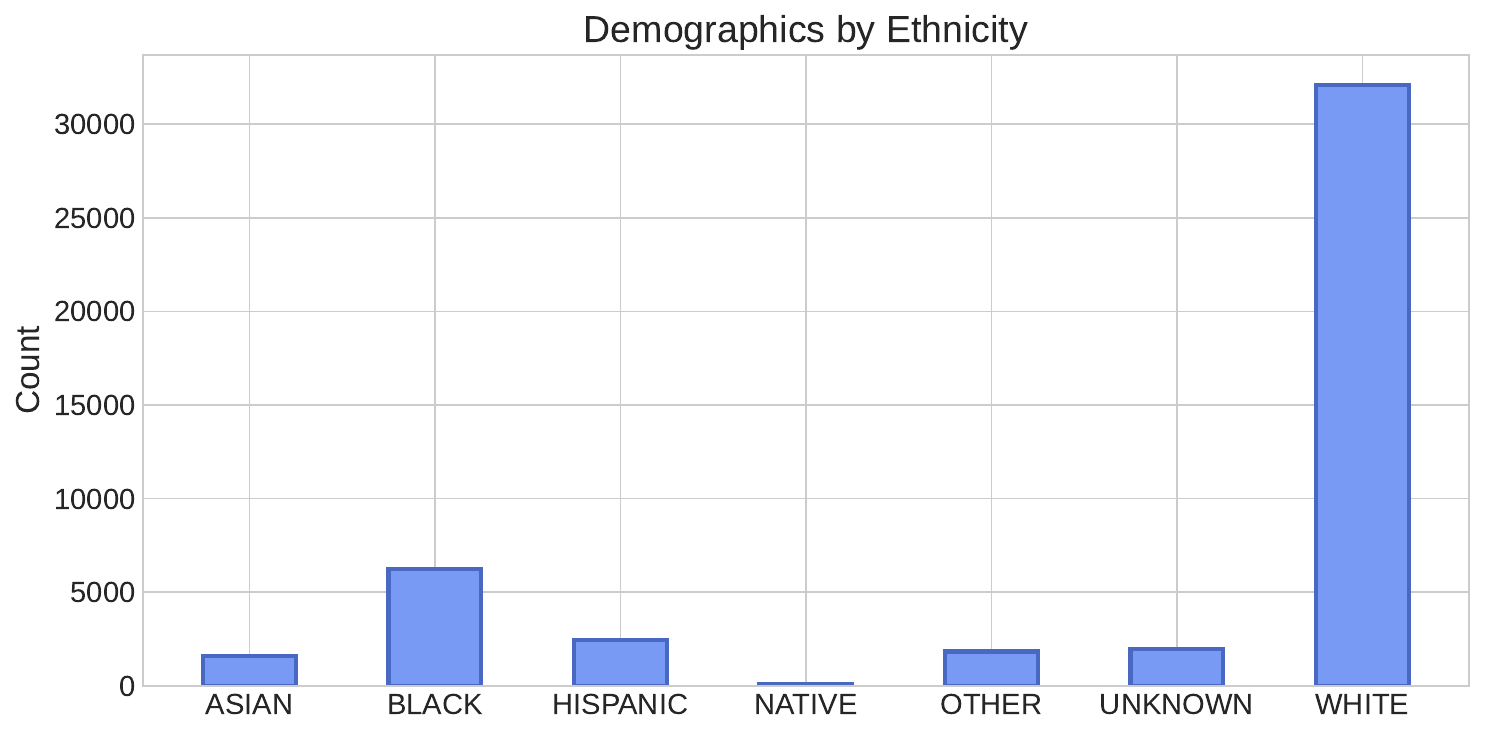}
        \subcaption{Ethnicity}
        \label{fig:mimiciv_demographics_ethnicity}
    \end{minipage}
\end{subfigure}
\hfill
\begin{subfigure}{1.0\linewidth}
    \begin{minipage}[c][][t]{1.0\linewidth}
        \includegraphics[width=1.0\textwidth]{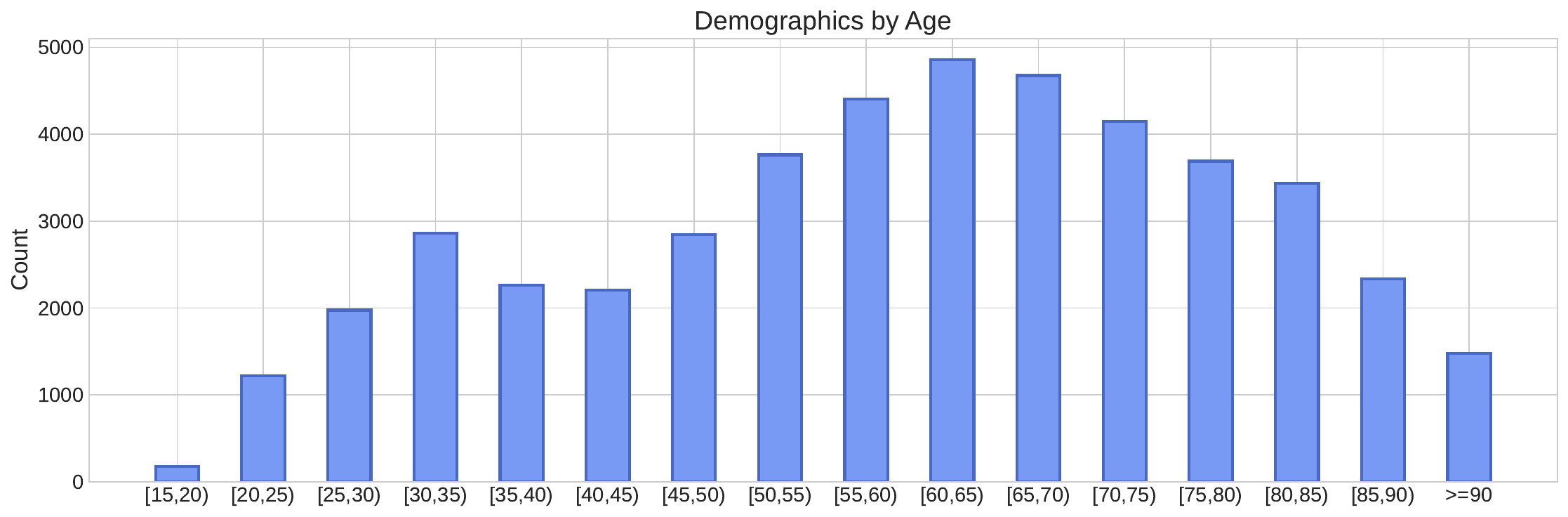}
        \subcaption{Age}
        \label{fig:mimiciv_demographics_age}
    \end{minipage}
\end{subfigure}
\caption{Summary of demographic bins in MIMIC-IV cohort.}
\label{fig:mimiciv_demographics_all}
\end{figure}

Overall data pre-processing for MIMIC-IV was carried out similarly to MIMIC-III as described previously. Since MIMIC-IV registered their clinical codes under both ICD-9 and ICD-10, we chose to filter and only utilized the samples with ICD-9-based diagnosis and procedure codes for this cohort. We present the statistics of the pre-processed cohorts of MIMIC-III and MIMIC-IV in Table \ref{table:preprocessed_cohorts} with augmented size referring to the total number of medical codes alongside their ancestor codes according to their respective ontology. Additionally, Table \ref{table:preprocessed_relations} outlines all extracted relations and their respective sizes, indicating the number of unique triples comprised of head and tail node types interconnected through a particular relation type in both cohorts. Since we regarded auxiliary demographic information (readily available in the cohort, such as gender, ethnicity, and age) to augment the medical KGs employing as a patient node, we presented a concise statistic summary of demographics using bar plots in Figure \ref{fig:demographics_all} and Figure \ref{fig:mimiciv_demographics_all} for of MIMIC-III and MIMIC-IV cohorts, respectively. 

For training and inference purposes, we split each cohort into the training set with a portion of two-thirds while equally allocating the remaining set for validation and testing. Furthermore, a bootstrap sampling \cite{yang2021safedrug} was adopted for evaluating the test set over ten rounds of repeated experiments (each with 80\% samples) to obtain recommended medications from the proposed and competing baselines.

\begin{table*}[!t]
\caption{Pre-processed cohort summary of MIMIC-III and MIMIC-IV cohorts.}
\label{table:preprocessed_cohorts}
\centering
\scalebox{1.0}{
    \begin{tabular}{lrrrr}
    \toprule
    \multicolumn{1}{c}{\multirow{2}{*}{\textbf{Items}}} & \multicolumn{2}{c}{\textbf{MIMIC-III}} & \multicolumn{2}{c}{\textbf{MIMIC-IV}}\\
    \cmidrule(lr){2-3} \cmidrule(lr){4-5}
     & \textbf{Original Size} & \textbf{Augmented Size} & \textbf{Original Size} & \textbf{Augmented Size} \\ \midrule
    Patients  & 6,350 & - & 46,332 & - \\
    Admissions & 15,032 & - & 111,489 & - \\
    Diagnoses & 1,958 & 3,051 & 2,000 & 3,207 \\
    Procedures & 1,428 & 1,975 & 1,984 & 2,604 \\
    Medicines & 113 & 188 & 128 & 208 \\ 
    Mean / Max of \# Admissions & 2.37 / 29 & - & 2.41 / 66 & - \\
    Mean / Max of \# Diagnoses over admissions & 10.51 / 128 & - & 7.99 / 197 & - \\
    Mean / Max of \# Procedures over admissions & 3.84 / 50 & - & 2.27 / 57 & - \\
    Mean / Max of \# Medicines over admissions & 11.65 / 64 & - & 7.71 / 72 & -\\  
    \bottomrule
    \end{tabular}
}
\end{table*}

\subsection{Implementation Settings} 
We initialized the dimensionality of node embeddings $\rmZ^{\{\rvc, \rvm, \rvs\}}_t$ to $64$ for each node that existed in the medical KGs $\gG^{*}$. For the R-GCN, we applied two layers followed by ReLU activation and a dropout of $0.5$ in the intermediate layer. The number of relations for R-GCN was set to $\{30,3\}$ based on the unique triple sets of clinical and medicine KGs, respectively. During training, we employed random edge-dropping as a graph augmentation technique. For the RNN, we utilized GRU cells for $\textrm{RNN}^{\{\rvc, \rvm, \rvc+\rvm \}}$ with hidden units of $64$. We fixed $\ell=3$ for the fusion module. Additionally, we set $\gH=4$ for APM, based on optimal empirical results from our preliminary ablation study, while setting $\lambda_{\textrm{rec}}=0.95$, $ \rho=0.05$ and $\tau=0.08$. All modules in KindMed were jointly trained using an Adam optimizer with a $2e-4$ learning rate and $1e-4$ weight decay over $200$ epochs. 

We implemented our KindMed\footnote{\href{https://anonymous.4open.science/r/KindMed-E28C/}{https://anonymous.4open.science/r/KindMed-E28C/}} using PyTorch 1.11.0. We further utilized PyTorch Geometric 2.2.0 to construct and learn the medical KGs using R-GCN layers. We conducted our exhaustive experiments on a Linux 18.04.5 LTS machine with 16 cores Intel Xeon Silver 4216 CPU, 8 $\times$ 32 GB memory, and 11GB NVIDIA RTX 2080 Ti GPU.

\subsection{Baselines} 
We designated our competing baselines by grouping them into two approaches: (i) \emph{instance-based}: Logistic Regression (LR), LEAP \cite{zhang2017leap}; (ii) \emph{longitudinal-based}: RETAIN \cite{choi2016retain}, GAMENet\cite{shang2019gamenet}, COGNet \cite{wu2022cognet}, SafeDrug \cite{yang2021safedrug}, Med-Tree \cite{yue2022medtree}. 

\begin{itemize}
    \item \textbf{LR} referred to logistic regression by treating the clinical multi-hot vector comprised of diagnosis and procedure code as the input feature. In practice, we employed LogisticRegression and OneVsRestClassifier method using scikit-learn 0.24.1 with its default configuration (\ie, L2 penalty, 100 maximum iterations, `LBFGS' solver).
    \item \textbf{LEAP \cite{zhang2017leap}} viewed a recommendation task as a process of sequence decision-making, predicting a set of recommended medicines per time step. We fixed the GRU-based feature encoder with 64 hidden units, a maximum length of 20 for generating the medications, and utilized the Adam optimizer with a learning rate of $2e-4$.
    \item \textbf{RETAIN \cite{choi2016retain}} sequentially predicted medication combination by exploiting two levels of attention mechanism, \ie, visit- and code-level features using GRUs. We fixed 64 hidden units for each employed GRU and trained it using Adam optimizer with a learning rate of $2e-4$.
    \item \textbf{GAMENet \cite{shang2019gamenet}} used graph-augmented memory networks by distilling the graphs of medication co-occurrences and DDIs using a late-fusion mechanism merged with the stored memory component for predicting the forthcoming medication combination. We employed two GRUs for aggregated diagnosis and procedure embeddings with 64 hidden units, respectively; two layers of independent GCNs, each with 64 hidden units, for both medication co-occurrences and DDI graphs; initialized 0.5 as the temperature $Temp$ with a decayed rate of $\epsilon=0.85$; 0.05 as its expected DDI rate; $\pi=[0.9, 0.1]$ as its mixture weight; and Adam optimizer with a learning rate of $2e-4$.
    \item \textbf{COGNet \cite{wu2022cognet}} utilized a copy-or-predict mechanism that considers the association between historical medication records and current diagnoses in recommending medicines. It distilled the medication embedding from the graph of medication co-occurrences and DDIs, realized through two layers of GCNs with 64 hidden units. We optimized COGNet using Adam optimizer with a $1e-3$ learning rate and 16 mini-batch sizes, $4$ beam size, and a maximum generation length of 45.
    \item \textbf{SafeDrug \cite{yang2021safedrug}} employed GNNs to learn global and local drug features by accounting for drugs' atomic bonds and chemical sub-structure co-existent in tackling the task of medication recommendation. We assigned a weight for BCE loss $\alpha=0.95$, a correcting factor $K_p = 0.05$, a Molecular Graph Neural Network with two layers, each comprised of 64 hidden units, and optimized SafeDrug using an Adam optimizer with a learning rate of $5e-4$.
    \item \textbf{Med-Tree \cite{yue2022medtree}} extended GAMENet by modifying attention-based modules for ontology-augmented graphs and sequence learning of diagnosis and procedure features. We followed the aforementioned overall settings of GAMENet with the following distinction: two separate GATs for learning the respective diagnosis and procedure ontology tree, each with two layers of 64 hidden units; separate GRUs for attention mechanism over aggregated diagnosis and procedure embeddings, each with 64 hidden units; a learning rate of $1e-4$.
\end{itemize}
We exploited the repository code of \cite{shang2019gamenet,wu2022cognet,yang2021safedrug} to implement these baselines. To ensure a fair comparison, we included ancestor codes for each observed clinical code in all baselines.

\begin{table}[!t]
\caption{Harnessed relations from external knowledge upon MIMIC-III and MIMIC-IV cohorts.}
\label{table:preprocessed_relations}
\centering
\scalebox{0.85}{
    \begin{tabular}{lcccrr}
    \toprule
    \multicolumn{1}{c}{\multirow{2}{*}{\textbf{Relations ($r$)}}} & \multicolumn{1}{c}{\multirow{2}{*}{\textbf{Type}}} & \multicolumn{1}{c}{\multirow{2}{*}{\textbf{Source}}} & \multicolumn{1}{c}{\multirow{2}{*}{\textbf{KG}}} & \multicolumn{2}{c}{\textbf{Total Size}} \\
    \cmidrule{5-6}
     &  &  &  & \textbf{MIMIC-III} &   \textbf{MIMIC-IV}\\ \midrule
    ICD-9 & Ont. & BioPortal & $\gG^{\rvc}$ & 5,022 & 5,807\\ 
    AFFECTS & Sem. & SemMedDB & $\gG^{\rvc}$ & 5,218 & 7,241 \\ 
    ASSOCIATED\_WITH & Sem. & SemMedDB & $\gG^{\rvc}$ & 1,851 & 3,007\\ 
    AUGMENTS & Sem. & SemMedDB & $\gG^{\rvc}$ & 827 & 1,009 \\ 
    CAUSES & Sem. & SemMedDB & $\gG^{\rvc}$ & 9,688 & 12,912 \\ 
    COEXISTS\_WITH & Sem. & SemMedDB & $\gG^{\rvc}$ & 18,216 & 25,857\\ 
    COMPLICATES & Sem. & SemMedDB & $\gG^{\rvc}$ & 2,521 & 3,083 \\ 
    DIAGNOSES & Sem. & SemMedDB & $\gG^{\rvc}$ & 1,601 & 2,627 \\ 
    ISA & Sem. & SemMedDB & $\gG^{\rvc}$ & 1,651 & 2,157 \\ \pagebreak
    MANIFESTATION\_OF & Sem. & SemMedDB & $\gG^{\rvc}$ & 1,160 & 1,600 \\ 
    MEASURES & Sem. & SemMedDB & $\gG^{\rvc}$ & 241 & 381 \\ 
    METHOD\_OF & Sem. & SemMedDB & $\gG^{\rvc}$ & 441 & 622 \\ 
    PRECEDES & Sem. & SemMedDB & $\gG^{\rvc}$ & 3,208 & 4,311 \\ 
    PREDISPOSES & Sem. & SemMedDB & $\gG^{\rvc}$ & 5,708 & 7,662 \\ 
    PREVENTS & Sem. & SemMedDB  & $\gG^{\rvc}$ & 585 & 785 \\ 
    TREATS & Sem. & SemMedDB & $\gG^{\rvc}$ & 3,453 & 5,165 \\ 
    ATC & Ont. & BioPortal & $\gG^{\rvm}$ & 174 & 194 \\ 
    DDI* & DDIs & TWOSIDES & $\gG^{\rvm}$ & 670 & 798 \\ 
    \bottomrule
    \multicolumn{5}{l}{{\small Ont.: Ontology; Sem.: Semantics; DDIs: Drug-Drug Interactions.
    }}
    \end{tabular}

}
\end{table}

\subsection{Evaluation Metrics} 
We evaluated the performance of all models in the medicine recommendation task using common metrics from the literature \cite{shang2019gamenet,yang2021safedrug}, including DDI rate ($\downarrow$), Jaccard score ($\uparrow$), precision-recall area under the curve (PRAUC) ($\uparrow$), F1-score ($\uparrow$), and the average number of prescribed medicines (Avg \# Meds). The arrow symbol ($\uparrow$) indicates that a higher value is favorable, while ($\downarrow$) indicates the opposite. 

\begin{itemize}
    \item \textbf{DDI rate ($\downarrow$).} We evaluated the DDI rate $\delta_{\textrm{DDI}}^n$ for a patient $n$ across their admission $T^n$ by aggregating the percentage of medicines that invoke DDIs as
    \begin{align*}
        \delta_{\textrm{DDI}}^n = \frac{\sum_t^{T^{n}} \sum_{i,j\in\{k:\hat{\rvm}_t^{n}[k]=1\}}  \mathbbm{1}\{\rmR_{\textrm{DDI}}[i,j]=1\}}{\sum_t^{T^{n}} \sum_{i,j\in\{k:\hat{\rvm}_t^{n}[k]=1\}} 1},
    \end{align*}
    where $\rmR_{\textrm{DDI}}\withdim{|\gM|\times|\gM|}$ denotes the DDI adjacency matrix, and $\mathbbm{1}$ denotes the indicator function (return one if the given condition matched or zero otherwise).

    \item \textbf{Jaccard ($\uparrow$).} We measured an average Jaccard score for a patient $n$ by evaluating the intersection of the recommended medicines with the ground truth and dividing it by their union.    
    \begin{align*}
        \textrm{Jaccard}_t^n &= \frac{|\{i: \rvm_t^{n}[i]=1\} \cap \{i: \hat{\rvm}_t^{n}[i]=1\}|}{|\{i: \rvm_t^{n}[i]=1\} \cup \{i: \hat{\rvm}_t^{n}[i]=1\}|}, \\
        \textrm{Jaccard}^n &= \frac{1}{T^n} \textstyle\sum_{t=1}^{T^{n}} \textrm{Jaccard}_t^n,
    \end{align*}
    \item \textbf{F1-score ($\uparrow$).} The average F1-score is calculated as the harmonic mean between precision and recall. For a given patient $n$, it can be measured as follows.
    \begin{align*}
        \textrm{Precision}^{n}_t &= \frac{|\{i: \rvm_t^{n}[i]=1\} \cap \{i: \hat{\rvm}_t^{n}[i]=1\}|}{|\ \{i: \hat{\rvm}_t^{n}[i]=1\}|} \\
        \textrm{Recall}^{n}_t &= \frac{|\{i: \rvm_t^{n}[i]=1\} \cap \{i: \hat{\rvm}_t^{n}[i]=1\}|}{|\{i: \rvm_t^{n}[i]=1\}|} \\
        \textrm{F1}_t^n &= \frac{2 \ \textrm{Precision}^{n}_t \ \textrm{Recall}^{n}_t}{\textrm{Precision}^{n}_t + \textrm{Recall}^{n}_t} \\
        \textrm{F1}^n &= \frac{1}{T^n} \textstyle\sum_{t=1}^{T^{n}} \textrm{F1}_t^n
    \end{align*}
    \item \textbf{Precision-Recall AUC (PRAUC) ($\uparrow$).} We evaluated the average PRAUC for a patient $n$ through
    \begin{align*}
        \Delta\textrm{Recall}(k)^n_t &= \textrm{Recall}(k)^{n}_t - \textrm{Recall}(k-1)^{n}_t \\
        \textrm{PRAUC}_t^n &= \sum_{i=1}^{|\gM|} \textrm{Precision}(k)^{n}_t \Delta\textrm{Recall}(k)^n_t, \\
        \textrm{PRAUC}^n &= \frac{1}{T^n} \textstyle\sum_{t=1}^{T^{n}} \textrm{PRAUC}_t^n,
    \end{align*}
    with $k$ referring to the rank over medicines, $\textrm{Precision}(k)^{n}_t$ as the precision with cut-off $k$ over the ordered list, and $\Delta\textrm{Recall}(k)^n_t$ denoting the changes.
\end{itemize}

\begin{table*}[!t]
\caption{Performances of KindMed on MIMIC-III based on number of heads $\gH$.}
\label{table:ablation_number_of_heads}
\centering
\scalebox{1.0}{
\begin{tabular}{rccccc}
\toprule
\scalebox{1.2}{$\gH$} & \textbf{DDI Rate ($\downarrow$)} & \textbf{Jaccard ($\uparrow$)} & \textbf{PRAUC ($\uparrow$)} & \textbf{F1-score ($\uparrow$)} & \textbf{Avg \#Meds} \\ \midrule
1 & \textbf{0.0628±0.0005} & 0.5401±0.0033 & 0.7840±0.0026 & 0.6929±0.0029 & 20.5364±0.1516 \\
2 & 0.0645±0.0007 & 0.5408±0.0032 & 0.7841±0.0028 & 0.6931±0.0029 & 20.4420±0.1630 \\ 
4 & \textbf{0.0628±0.0006} & \textbf{0.5427±0.0033} & 0.7843±0.0026 & \textbf{0.6950±0.0029} & 20.2654±0.1736 \\ 
8 & 0.0650±0.0006 & 0.5402±0.0037 & \textbf{0.7853±0.0024} & 0.6926±0.0033 & 20.8139±0.1572 \\
16 & 0.0649±0.0005 & 0.5410±0.0031 & 0.7839±0.0025 & 0.6933±0.0028 & 20.9254±0.1560 \\
32 & 0.0666±0.0007 & 0.5415±0.0031 & 0.7845±0.0025 & 0.6939±0.0027 & 20.4798±0.1450 \\ 
\bottomrule
\end{tabular}
}
\end{table*}

\begin{table*}[!t]
\centering
\caption{Performances of KindMed variants on MIMIC-III by ablating its inputs, internal modules, and harnessed relations.}
\label{table:kindmed_ablation_studies}
\scalebox{0.9}{
\begin{tabular}{cccccccccccccc}
\toprule 
\multirow{2}{*}{\textbf{Model Variant}} & \multicolumn{5}{c}{\textbf{Input/Modules}} & \multicolumn{3}{c}{\textbf{Relations}} & \multirow{2}{*}{\textbf{DDI Rate ($\downarrow$)}} & \multirow{2}{*}{\textbf{Jaccard ($\uparrow$)}} & \multirow{2}{*}{\textbf{PRAUC ($\uparrow$)}} & \multirow{2}{*}{\textbf{F1-score ($\uparrow$)}} & \multirow{2}{*}{\textbf{Avg \#Meds}} \\ 
\cmidrule(lr){2-6} \cmidrule(lr){7-9}
& $\gG^{\rvc}$ & $\gG^{\rvm}$ & $\rvs$ & \textbf{F} & \textbf{AP} 
& \textbf{Ont.} & \textbf{Sem.} & \textbf{DDIs} \\
\midrule
\multicolumn{1}{l}{KindMed$_{\gG^{\rvm-}}$} & \CIRCLE & \Circle & \CIRCLE & \Circle & \CIRCLE & \CIRCLE & \CIRCLE & \Circle & 0.0637±0.0007 & 0.5353±0.0032 & 0.7800±0.0027 & 0.6884±0.0029 & 20.4331±0.1431 \\ 
\multicolumn{1}{l}{KindMed$_{\gG^{\rvm},\textrm{AP}^{-}}$} & \CIRCLE & \Circle & \CIRCLE & \Circle & \Circle & \CIRCLE & \CIRCLE & \Circle & 0.0659±0.0003 & 0.5334±0.0029 & 0.7794±0.0023 & 0.6869±0.0026 & 22.0806±0.2000 \\
\multicolumn{1}{l}{KindMed$_{\rvs^{-}}$} & \CIRCLE & \CIRCLE & \Circle & \CIRCLE & \CIRCLE & \CIRCLE & \CIRCLE & \CIRCLE & 0.0644±0.0004 & 0.5375±0.0032 &0.7844±0.0024 & 0.6902±0.0029 & 20.9893±0.1902 \\ 
\multicolumn{1}{l}{KindMed$_{\textrm{F}^{-}}$} & \CIRCLE & \CIRCLE & \CIRCLE & \Circle & \CIRCLE & \CIRCLE & \CIRCLE & \CIRCLE & 0.0649±0.0005 & 0.5397±0.0033 & \textbf{0.7851±0.0026} & 0.6923±0.0030 & 20.2222±0.1452 \\ 
\multicolumn{1}{l}{KindMed$_{\textrm{AP}^{-}}$} & \CIRCLE & \CIRCLE & \CIRCLE & \CIRCLE & \Circle & \CIRCLE & \CIRCLE & \CIRCLE & 0.0665±0.0004& 0.5400±0.0030 & 0.7840±0.0019 & 0.6925±0.0027 & 20.8084±0.1666 \\ 
\midrule \multicolumn{1}{l}{KindMed$_{\textrm{Rels.}^{-}}$} & \Circle & \Circle & \Circle & \CIRCLE &  \CIRCLE & \Circle & \Circle & \Circle & 0.0679±0.0006 & 0.5372±0.0022 & 0.7845±0.0024 &0.6899±0.0020 & 21.2790±0.1440 \\ 

\multicolumn{1}{l}{KindMed$_{\textrm{Ont.}^{-}}$} & \CIRCLE & \CIRCLE & \CIRCLE & \CIRCLE & \CIRCLE & \Circle & \CIRCLE & \CIRCLE & 0.0638±0.0004 & 0.5388±0.0030 & 0.7837±0.0024 & 0.6916±0.0027 & 21.0301±0.1768 \\ 
\multicolumn{1}{l}{KindMed$_{\textrm{Sem.}^{-}}$} & \CIRCLE & \CIRCLE & \CIRCLE & \CIRCLE & \CIRCLE & \CIRCLE & \Circle & \CIRCLE & 0.0679±0.0004 & 0.5424±0.0034 & 0.7844±0.0027 & 0.6945±0.0030 & 21.3285±0.1783 \\
\multicolumn{1}{l}{KindMed$_{\textrm{DDI}^{-}}$} & \CIRCLE & \CIRCLE & \CIRCLE & \CIRCLE & \CIRCLE & \CIRCLE & \CIRCLE & \Circle &  0.0661±0.0006 & 0.5402±0.0030 & 0.7834±0.0023 & 0.6930±0.0025 & 20.4546±0.1675 \\ \midrule
\multicolumn{1}{l}{KindMed$_{\textrm{Hist.}}$} & \CIRCLE & \CIRCLE & \CIRCLE & \CIRCLE & \CIRCLE & \CIRCLE & \CIRCLE & \CIRCLE & 0.0644±0.0005 & 0.5418±0.0036 & 0.7849±0.0027 & 0.6943±0.0033 & 20.7479±0.1600 \\ 
\multicolumn{1}{l}{KindMed} & \CIRCLE & \CIRCLE & \CIRCLE & \CIRCLE & \CIRCLE & \CIRCLE & \CIRCLE & \CIRCLE & \textbf{0.0628±0.0006} & \textbf{0.5427±0.0033} & 0.7843±0.0026 & \textbf{0.6950±0.0029} & 20.2654±0.1736 \\ 


\bottomrule
\end{tabular}
}\\
{\small 
\CIRCLE: w/; \Circle: wo/; $\gG^{\rvc}$: Clinical KG; $\gG^{\rvm}$: Medicine KG; $\rvs$: Demographic Features; F: Fusion; AP: Attentive Prescribing
}
\end{table*}


\subsection{Ablation Studies} 
We conducted an exhaustive ablation study on KindMed using the MIMIC-III cohort to investigate the influential roles of the number of heads in APM, inputs, internal modules, and relations, as presented in Table \ref{table:ablation_number_of_heads} and Table \ref{table:kindmed_ablation_studies}. We examined the optimal number of heads $\gH$ of the multi-head attention mechanism in APM by fixing the DDI threshold $\tau=0.08$ (based on the test set DDI rate) and experimented with $\gH = [1,2,4,8,16,32]$. The results showed that increasing the number of heads $\gH>4$ did not increase the overall performance. Thus, we fixed $\gH=4$ for the remaining experiments in this study, as they showed good overall outcomes.

KindMed$_{\gG^{\rvm-}}$ excluded the medical KG $\gG^{\rvm}_{1:t-1}$, using only the clinical KG $\gG^{\rvc}_{1:t}$ as the input graph. Consequently, we also removed the related modules, \ie, $\textrm{GNN}^{\rvm}$, $\textrm{RNN}^{\{\rvm, \rvc+\rvm\}}$, and the fusion module. At the last admission, we set the temporal clinical features $\rvh^{\rvc}_t$ as the APM inputs and applied an additive residual connection to itself. The absence of historical medication records led to lower prescription accuracies for this variant. By further removing APM and directly feeding $\rvh^{\rvc}_t$ to the predictive FFN, KindMed$_{\gG^{\rvm},\textrm{AP}^{-}}$ exhibited even poorer results. In KindMed$_{\rvs^{-}}$, we used both medical KGs $\gG^*_t$ but detached the demographic features. It performed better than the former ones, suggesting that personalized historical medical records alone can achieve decent performance even without demographic features. For KindMed$_{\textrm{F}^{-}}$, we replaced the fusion module with a simple concatenation of the learned temporal features, \ie, $\rvf_{t-1} = [\rvh^{\rvc}_{t-1} \circ \rvh^{\rvm}_{t-1}]$. This variant showed considerably higher accuracy and better medication safety performance than prior variants, indicating that merely combining both learned temporal patient features was sufficient to recommend precise medicines. KindMed$_{\textrm{AP}^{-}}$ exchanged the beneficial MHA in APM with an FFN to merge the temporal features, \ie,  $\rvo_t = \textrm{LNorm}(\rvg^{\rvc}_t + \textrm{FFN}(\rvh^{\rvc}_t,\ \rvh^{\rvc+\rvm}_{t-1}))$. While it exhibited overall good performances, it resulted in a higher DDI rate compared to the prior variants. All of these variants performed inferior to the fully functional KindMed in most evaluation metrics, underscoring the importance of utilizing all inputs/modules to achieve the best recommendation outcomes.

We further examined KindMed$_{\textrm{Rels.}^{-}}$, which ignored all types of relations, treating the medical codes as independent inputs. For this, we aggregated the embedding of clinical and medicine codes to be directly fed into hierarchical sequence learning. We retained the remaining modules, \ie, fusion and APM modules, unchanged. This variant struggled to achieve satisfactory recommendation accuracy, and its DDI rate was among the highest of all variants. This outcome highlighted the significant disadvantages of neglecting relations between medical codes. KindMed$_{\textrm{Ont.}^{-}}$ excluded ICD and ATC as its medical ontology embodied in medical KGs $\gG^{*}_t$, achieving a decent DDI rate but at the cost of accuracy. KindMed$_{\textrm{Sem.}^{-}}$ pruned a wide range of semantic relations between clinical nodes in $\gG^\rvc_t$, relying solely on the hierarchical linkages of ICD ontology between clinical codes. This variant showed improved results among all variants, suggesting that utilizing ontology enhances recommendation outcomes. However, it was more prone to a higher risk of DDI. Meanwhile, KindMed$_{\textrm{DDI}^{-}}$ dropped the DDI relations among medicine nodes in $\gG^\rvm_t$, leading to a higher DDI risk. This highlighted the importance of integrating such relations into medicine KG. In contrast, the variant of KindMed that employed all relations performed better than the preceding relation-ablated variants. 

Finally, instead of only exploiting the penultimate temporal summary $\rvh^{\rvc+\rvm}_{t-1}$, we could incorporate the entire set of temporal features $\rvh^{\rvc+\rvm}_{1:t-1}$ (\ie, by concatenating features across all timesteps) to be injected into MHA. This approach allowed the attended features to consider all preceding admission records attentively. While KindMed utilized the former, KindMed$_{\textrm{Hist.}}$ employed the latter. The outcomes between these two variants were fairly comparable, with slightly upper-hand results favoring KindMed. This suggests that utilizing a penultimate joint summary of temporal features is sufficient to secure satisfactory medication recommendations.

\begin{table*}[!t]
\centering
\caption{Performance comparison of KindMed against several baselines in recommending medicines. $*: p$-value $<0.05$}
\label{table:kindmed_vs_baselines}
\scalebox{1.0}{
\begin{tabular}{clccccc}
\toprule
\multicolumn{1}{c}{\textbf{Dataset}} & \multicolumn{1}{c}{\textbf{Model}} & \textbf{DDI Rate ($\downarrow$)} & \textbf{Jaccard ($\uparrow$)} & \textbf{PRAUC ($\uparrow$)} & \textbf{F1-score ($\uparrow$)} & \textbf{Avg \#Meds} \\ \midrule
\multirow{10}{*}{MIMIC-III} & Ground Truth & 0.0753±0.0010 & - & - & - & 20.3564±0.1341 \\ \cmidrule{2-7}
& LR & 0.0779±0.0011$^{*}$ & 0.4830±0.0017$^{*}$ & 0.7477±0.0018$^{*}$ & 0.6423±0.0016$^{*}$ & 17.7562±0.0803 \\ 
& LEAP \cite{zhang2017leap} & 0.0708±0.0008$^{*}$ & 0.4539±0.0022$^{*}$ & 0.6680±0.0029$^{*}$ & 0.6167±0.0022$^{*}$ & 18.6301±0.0607 \\ \cmidrule{2-7}
& RETAIN \cite{choi2016retain} & 0.0823±0.0009$^{*}$ & 0.4803±0.0040$^{*}$ & 0.7567±0.0041$^{*}$ & 0.6418±0.0039$^{*}$ & 17.7431±0.1814 \\ 
& GAMENet \cite{shang2019gamenet} & 0.0764±0.0006$^{*}$ & 0.5219±0.0024$^{*}$ & 0.7711±0.0029$^{*}$ & 0.6767±0.0022$^{*}$ & 24.3748±0.1291 \\ 
& COGNet \cite{wu2022cognet} & 0.0830±0.0006$^{*}$ & 0.5171±0.0022$^{*}$ & 0.7683±0.0030$^{*}$ & 0.6729±0.0021$^{*}$ & 31.3660±0.2823 \\
& SafeDrug \cite{yang2021safedrug} & \underline{0.0656±0.0006$^{*}$} & 0.5227±0.0025$^{*}$ & \underline{0.7743±0.0027$^{*}$} & 0.6774±0.0022$^{*}$ & 21.8138±0.1215 \\
& Med-Tree \cite{yue2022medtree} & 0.0800±0.0006$^{*}$ & \underline{0.5270±0.0019$^{*}$} & 0.7741±0.0025$^{*}$ & \underline{0.6809±0.0018$^{*}$} & 24.3199±0.1704 \\
\cmidrule{2-7}
& \textbf{KindMed (Ours)} & \textbf{0.0628±0.0006} & \textbf{0.5427±0.0033} & \textbf{0.7843±0.0026} & \textbf{0.6950±0.0029} & 20.2654±0.1736 \\
\midrule
\multirow{10}{*}{MIMIC-IV} & Ground Truth & 0.0862±0.0004 & - & - & - & 13.7821±0.0540 \\ 
\cmidrule{2-7}
& LR & 0.0775±0.0004 & 0.4837±0.0006 & 0.7503±0.0006 & 0.6353±0.0005 & 10.9786±0.0312 \\ 
& LEAP \cite{zhang2017leap} & 0.0841±0.0004$^{*}$ & 0.4733±0.0024$^{*}$ & 0.6444±0.0018$^{*}$ & 0.6274±0.0022$^{*}$ & 14.4765±0.0430 \\ 
\cmidrule{2-7}
& RETAIN \cite{choi2016retain} & 0.0904±0.0011$^{*}$ & 0.4613±0.0026$^{*}$ & 0.7218±0.0018$^{*}$ & 0.6170±0.0023$^{*}$ & 12.8949±0.0923 \\ 
& GAMENet \cite{shang2019gamenet} & 0.0848±0.0005$^{*}$ & 0.4920±0.0018$^{*}$ & 0.7487±0.0015$^{*}$ & 0.6449±0.0017$^{*}$ & 19.3289±0.0912 \\ 
& COGNet \cite{wu2022cognet} & 0.0956±0.0004$^{*}$ & 0.4933±0.0017$^{*}$ & \underline{0.7525±0.0018}$^{*}$ & 0.6471±0.0017$^{*}$ & 23.6343±0.1121 \\ 
& SafeDrug \cite{yang2021safedrug} & \underline{0.0737±0.0004}$^{*}$ & \underline{0.5065±0.0020}$^{*}$ & 0.7503±0.0013$^{*}$ & \underline{0.6578±0.0019}$^{*}$ & 15.9642±0.0335 \\ 
& Med-Tree \cite{yue2022medtree} & 0.0859±0.0006$^{*}$ & 0.5008±0.0017$^{*}$ & \textbf{0.7552±0.0017}$^{*}$ & 0.6531±0.0017$^{*}$ & 18.7438±0.0756 \\ 
\cmidrule{2-7}
& \textbf{KindMed (Ours)} & \textbf{0.0684±0.0003} & \textbf{0.5073±0.0016} & 0.7517±0.0014 & \textbf{0.6586±0.0016} & 15.4589±0.0361\\  \bottomrule
\end{tabular}
}
\end{table*}

\subsection{Performances Against Baselines}  
We validated the effectiveness of our proposed model by reporting the performance comparison of our KindMed model against several competing baselines on MIMIC-III in Table \ref{table:kindmed_vs_baselines}. We included the ground truth of medications over the repeated evaluation on the test set. As such, we presented our KindMed performance with a DDI threshold $\tau=0.08$ to restrain the DDI rate accordingly. The results showed that instance-based approaches, such as LR and LEAP, performed inferior to their longitudinal counterparts. Among the longitudinal-based approaches, GAMENet, SafeDrug, and Med-Tree emerged as the top-tier achieving baselines. These models were reinforced with effective objective functions designed to constrain the DDI rate and promote medication safety in recommending medicines. Med-Tree, in particular, extended GAMENet with attention mechanisms for ontology-augmented graph representation and sequence learning, which increased the overall metrics but resulted in a higher DDI rate as a trade-off. Our proposed KindMed model outperformed all designated baselines in overall evaluation metrics. The average number of prescribed medicines by KindMed was comparable to the ground truth, but it was more effective in suppressing the DDI rate while maintaining greater recommendation accuracies. This showcased the capability of KindMed to provide safe and accurate medication recommendations.

For the evaluation on MIMIC-IV, we reported our KindMed with a fixed $\tau=0.09$. We adjusted the training epochs of the models to 100 and set the learning rate accordingly. In this particular EHR cohort, our KindMed performed with comparable or better accuracy and a lower DDI rate compared to the baselines. Given that the MIMIC-IV cohort contained more extensive records and a wider range of medical codes, we can firmly observe that our proposed KindMed is well-equipped to handle the medicine recommendation task across diverse real-world EHR cohorts.


\subsection{Controllable DDI Rate Thresholds} 
\begin{figure}[!t]
  \centering
  \includegraphics[width=1.0\linewidth]{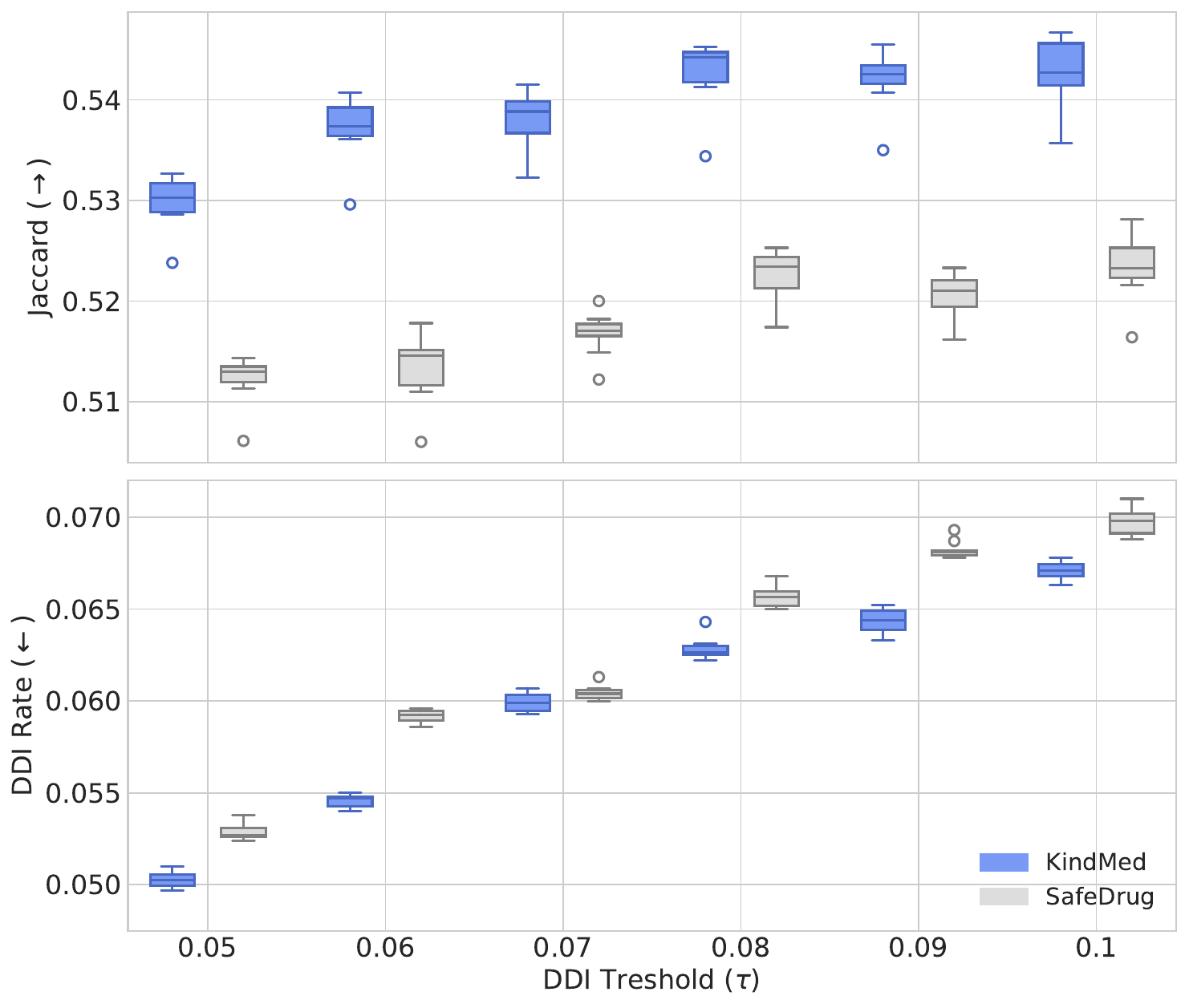}
  \caption{Performance comparison of KindMed and SafeDrug by varying DDI threshold hyperparameter.}
  \label{fig:kindmed_vs_safedrug}
\end{figure}

Given our paramount concern for medication safety, we conducted an in-depth analysis of our proposed KindMed model's ability to restrain the DDI rate through the DDI threshold hyperparameter $\tau$. Specifically, we directly compared our model's performances with SafeDrug \cite{yang2021safedrug} using the MIMIC-III cohort, as both models employed equivalent objective functions to optimize the entire model. We set the DDI threshold within the range of $\left[0.05, 0.1\right]$, with an increment of 0.01. The resulting performances in terms of Jaccard and DDI rate are illustrated in Figure \ref{fig:kindmed_vs_safedrug}. Notably, our KindMed consistently managed to achieve lower DDI rates across varying thresholds while maintaining a higher Jaccard score compared to SafeDrug.


\subsection{Medication Scenario \& Error Analysis}
\label{medication_scenarios}
In order to understand the behavior of our KindMed in recommending medications, we performed additional analysis by means of medication scenarios. For this purpose, we reported a couple of case studies of medication, namely good and poor medication cases, against top-performing baselines on the MIMIC-III cohort in Table \ref{table:medication_scenario_good} and Table \ref{table:medication_scenario_poor}, respectively. To accomplish this, we ordered the averaged F1 scores obtained by our recommendation results over test samples, as suggested by the prior study of \cite{sun2022drugrec}. We observed that our KindMed exhibited recommending medicine combinations that promoted fewer false-positive cases in both scenarios.

\begin{table}[!t]
\centering
\caption{Good medication case of KindMed against baselines on MIMIC-III cohort.}
\label{table:medication_scenario_good}
\scalebox{0.77}{
\begin{tabular}{lllcc}
\toprule
& \multicolumn{1}{c}{\textbf{Model}} & \multicolumn{1}{c}{\textbf{Medical Codes}} & \textbf{F1-Score} & \textbf{DDI Rate} \\ \hline
\multirow{6}{*}{\rotatebox[origin=c]{90}{1st Visit}} & GT & \makecell[l]{\textbf{Diagnosis:} 348.89, 401.9, 518.0, 780.39 \\
\textbf{Procedures:} 96.71 \\
\textbf{Medicines:} A01A, A02B, A04A, A06A, \\ A12B, A12C, B01A, B05C, C02D, C03A, \\ H04A, N01A, N02B, N03A} & - & 0.0220 \\ \cmidrule(lr){2-5}
& GAMENet & \makecell[l]{\textbf{Medicines:} 12 Correct + 2 Missed + 5 Unseen} & 0.7742 & 0.0735 \\ \cmidrule(lr){2-5}
& SafeDrug & \makecell[l]{\textbf{Medicines:} 11 Correct + 3 Missed + 5 Unseen} & 0.7333 & 0.0750 \\ \cmidrule(lr){2-5} 
& Med-Tree & \makecell[l]{\textbf{Medicines:} 11 Correct + 3 Missed + 2 Unseen} & 0.8148 & 0.0256 \\ \cmidrule(lr){2-5} 
& KindMed & \makecell[l]{\textbf{Medicines:} 10 Correct + 4 Missed} & 0.8333 & 0.0222 \\ \midrule
\multirow{7}{*}{\rotatebox[origin=c]{90}{2nd Visit}} & GT & \makecell[l]{\textbf{Diagnosis:} 191.2, 401.9 \\
\textbf{Procedures:} 00.39, 01.59, 02.12 \\
\textbf{Medicines:} A01A, A02B, A04A, A06A, \\ A12C, B01A, B05C, C02D,  C03A, C07A, \\ J01D, N02A, N02B, N03A} & - & 0.0659 \\ \cmidrule(lr){2-5}
& GAMENet & \makecell[l]{\textbf{Medicines:} 13 Correct + 1 Missed + 3 Unseen} & 0.8667 & 0.0417 \\ \cmidrule(lr){2-5}
& SafeDrug & \makecell[l]{\textbf{Medicines:} 13 Correct + 1 Missed + 5 Unseen} & 0.8125 & 0.0588 \\ \cmidrule(lr){2-5}
& Med-Tree & \makecell[l]{\textbf{Medicines:} 13 Correct + 1 Missed + 4 Unseen} & 0.8387 & 0.0441 \\ \cmidrule(lr){2-5} 
& KindMed & \makecell[l]{\textbf{Medicines:} 13 Correct + 1 Missed + 1 Unseen} & 0.9286 & 0.0549 \\ 
\bottomrule
\end{tabular}
}
\end{table}

\begin{figure*}[!t]
\centering
\begin{subfigure}{0.32\textwidth}
    \begin{minipage}[b][][t]{1.0\linewidth}
        \includegraphics[width=1.0\textwidth]{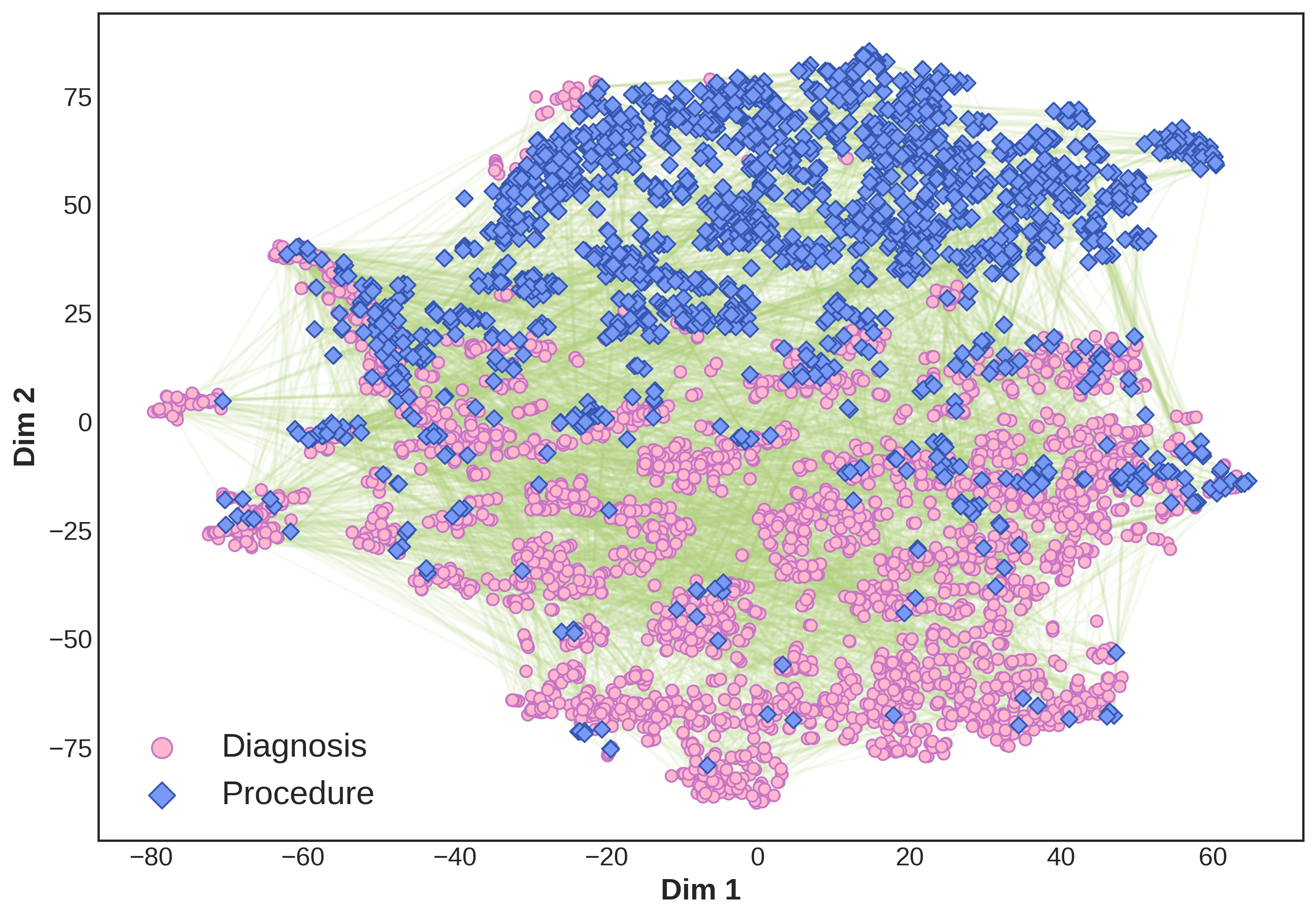}
        \subcaption{KindMed$_{\textrm{Sem.}^{-}}$}
        \label{fig:tsne_visualization_kindmed_wo_semantics}
    \end{minipage}
\end{subfigure}
\hfill
\begin{subfigure}{0.32\textwidth}
    \begin{minipage}[b][][t]{1.0\linewidth}
        \includegraphics[width=1.0\textwidth]{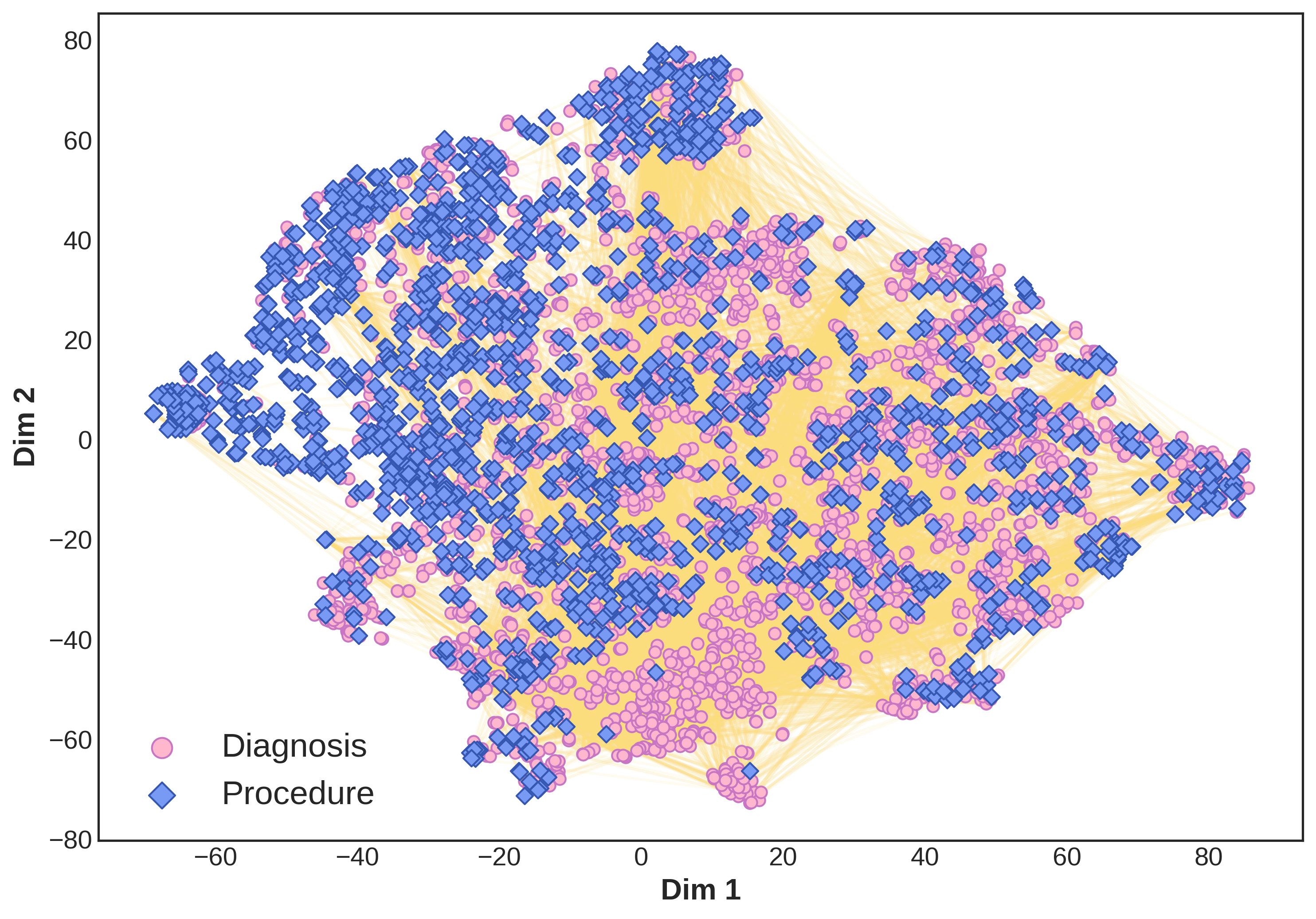}
        \subcaption{KindMed$_{\textrm{Ont.}^{-}}$}
        \label{fig:tsne_visualization_kindmed_wo_ontology}
    \end{minipage}
\end{subfigure}
\hfill
\begin{subfigure}{0.32\textwidth}
    \begin{minipage}[b][][t]{1.0\linewidth}
        \includegraphics[width=1.0\textwidth]{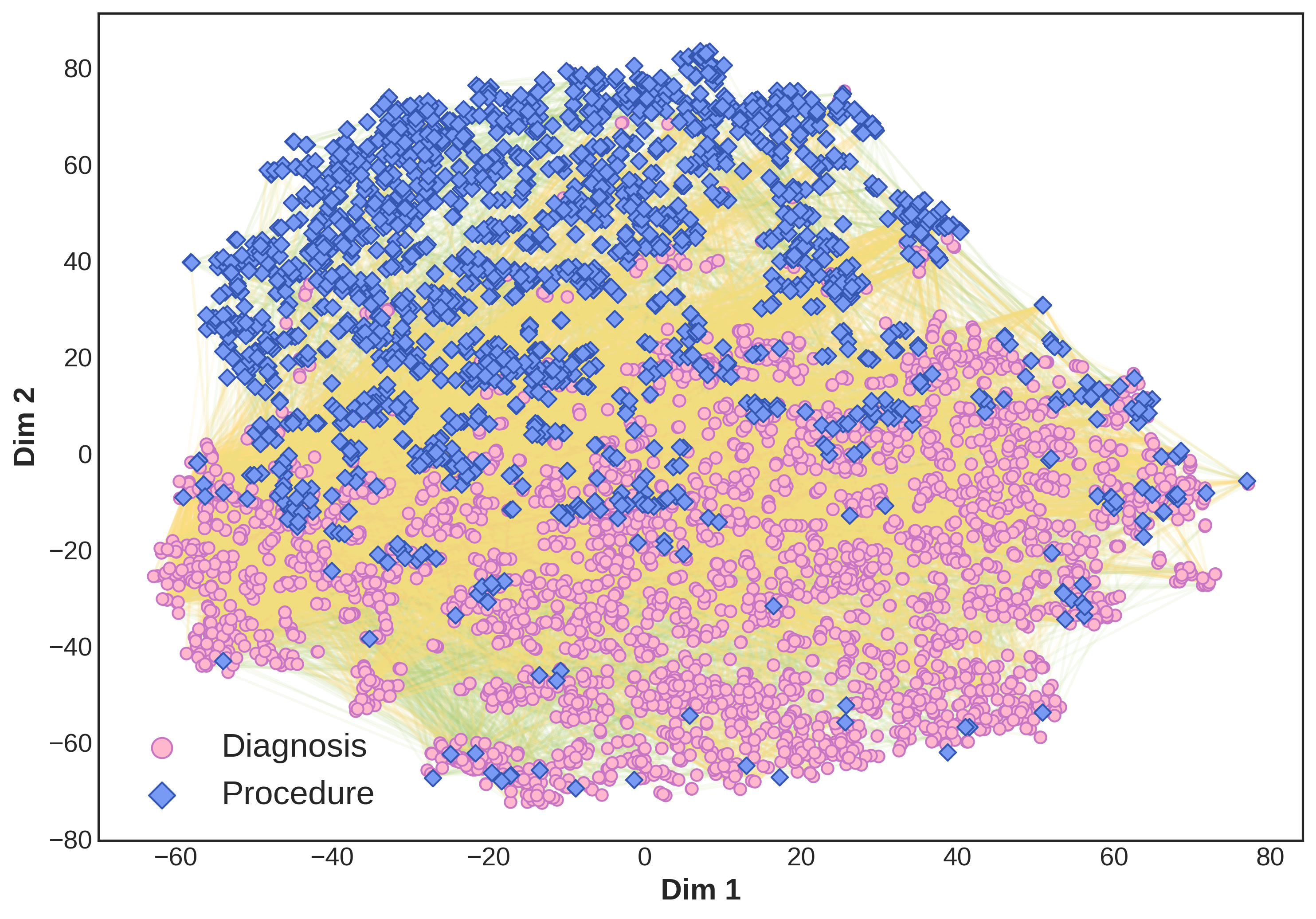}
        \subcaption{KindMed}
        \label{fig:tsne_visualization_kindmed_full}
    \end{minipage}
\end{subfigure}
\caption{The t-SNE plots over KindMed variants on the MIMIC-III cohort. Ontology and semantic relations are shown in green and yellow, respectively. We deliberately showed semantic inter-relations across two entities to make them more prominent. }
\label{fig:tsne_visualization}
\end{figure*}

\begin{table}[!t]
\centering
\caption{Poor medication case of KindMed against baselines on MIMIC-III cohort.}
\label{table:medication_scenario_poor}
\scalebox{0.71}{
\begin{tabular}{lllcc}
\toprule
& \multicolumn{1}{c}{\textbf{Model}} &  \multicolumn{1}{c}{\textbf{Medical Codes}} & \textbf{F1-Score} & \textbf{DDI Rate} \\ \hline
\multirow{7}{*}{\rotatebox[origin=c]{90}{1st Visit}} & GT & \makecell[l]{\textbf{Diagnosis:} 250.12, 250.22, 250.42, 272.4,  \\ 274.9,  276.1, 276.7, 278.00, 287.5, 293.0, \\ 300.00, 327.23, 403.90, 414.01, 426.0, 427.5, \\ 453.81, 455.6, 459.81, 507.0, 584.9, 585.3, \\ 785.51, 996.74,  E879.8, V15.81, V16.0, V58.67 \\
\textbf{Procedures:} 31.42, 38.91, 38.93, 96.04, 96.71, 99.69 \\
\textbf{Medicines:} A01A, A02B, A06A, A10A, \\ A10B, A11D, A12B, A12C,  B01A, B03B, \\ B05C, C01E,  C02D, C03A, C07A, C09A, \\ C10A, H04A, J01D, J01F, N01A, N02A, \\N02B, N05A, N05B, N05C} & - & 0.0800 \\ \cmidrule(lr){2-5}
& GAMENet & \makecell[l]{\textbf{Medicines:} 19 Correct + 7 Missed + 12 Unseen} & 0.6667 & 0.0688 \\ \cmidrule(lr){2-5}
& SafeDrug & \makecell[l]{\textbf{Medicines:} 18 Correct + 8 Missed + 11 Unseen} & 0.6545 & 0.0788 \\ \cmidrule(lr){2-5} 
& Med-Tree & \makecell[l]{\textbf{Medicines:} 17 Correct + 9 Missed + 12 Unseen} & 0.6182 & 0.0714 \\ \cmidrule(lr){2-5} 
& KindMed & \makecell[l]{\textbf{Medicines:} 16 Correct + 10 Missed + 7 Unseen} & 0.6531 & 0.0395 \\ \midrule
\multirow{6}{*}{\rotatebox[origin=c]{90}{2nd Visit}} & GT & \makecell[l]{\textbf{Diagnosis:} 250.40, 272.4, 274.9, 278.00,  300.00, \\ 327.23, 403.90, 414.01, 414.2, 426.4,  427.5, 443.9,  \\ 518.82,  585.9, 785.51, V15.82, V49.87, V58.67 \\
\textbf{Procedures:} 37.21, 37.61, 37.78, 88.56, 96.71, 99.60 \\
\textbf{Medicines:} A01A, A02B} & - & 0.0000 \\ \cmidrule(lr){2-5}
& GAMENet & \makecell[l]{\textbf{Medicines:} 2 Correct + 26 Unseen} & 0.1333 & 0.0979 \\ \cmidrule(lr){2-5}
& SafeDrug & \makecell[l]{\textbf{Medicines:} 2 Correct + 26 Unseen} & 0.1333 & 0.0847 \\ \cmidrule(lr){2-5}
& Med-Tree & \makecell[l]{\textbf{Medicines:} 2 Correct + 25 Unseen} &  0.1379 & 0.0997 \\ \cmidrule(lr){2-5} 
& KindMed & \makecell[l]{\textbf{Medicines:} 2 Correct + 15 Unseen} & 0.2105 & 0.0588 \\ 
\bottomrule
\end{tabular}
}
\end{table}

\subsection{Knowledge-Induced Node Embeddings} 
To delve into the influence of relations on embedding spaces, we analyzed the node embeddings discovered by relation-aware GNNs. We exploited t-SNE \cite{maaten2008tsne} to visualize the node embedding of clinical KGs $\gG^{\rvc}$ aggregated over all patients over the MIMIC-III cohort, as depicted in Figure \ref{fig:tsne_visualization}. Our KindMed$_{\textrm{Sem.}^{-}}$ variant, which disregarded semantic relations (thus, it relied on ontology as the hierarchical structures), exhibited distinct regions of diagnoses and procedures within the embedding space. Sub-clusters within each entity type were also discernible. Conversely, KindMed$_{\textrm{Ont.}^{-}}$ produced sparser and more intertwined embedding of nodes across different entity types. Since the semantic relations pulled clinical nodes closer semantically, this pattern was exposed by the entanglement between such two entities. The full KindMed in Figure \ref{fig:tsne_visualization_kindmed_full} showcased relatively clear boundaries between diagnoses and procedures, with semantic entanglement observed in intersection spaces while forming finer granularity spreads, highlighting a more comprehensive relational representation.

\section{Discussion} 
\label{section:discussions}

According to the reported experimental results, our proposed KindMed outlined a positive research direction in the avenue of graph-driven medicine recommendation. Nevertheless, we have reflected on several points worthy of further discussion to highlight certain limitations and explore future directions. Notably, the utilization of KGs poses inherent challenges, including noise and incompleteness \cite{shi2018open}. While we relaxed this issue by performing random edge-dropping as a graph augmentation, there are potential approaches for improvement. One approach worth considering is pre-training the relation-aware GNNs on pretext tasks, which could enhance and encourage more expressive KG embeddings. 

Furthermore, the current study prioritized recommendation accuracy and medication safety over explainability. An in-depth analysis and extension to the proposed framework shall be carried out to account for the human-interpretable rationale behind prescribed medications tailored to the patient's health condition. For instance, by dynamically attending a subset of crucial nodes or connections in medical KGs across admissions that strongly contribute to medication recommendation.

Further investigation into different types of medical codes under other well-standardized ontologies (\eg, Systematized Nomenclature of Medicine-Clinical Terms (SNOMED-CT), multi-level Clinical Classifications Software (CCS)) needs to be conducted rigorously to observe their structural/hierarchical influences on enriching the node embeddings. Additionally, extracting more diverse semantic relations between clinical and medicine codes is feasible with the assistance of large language models (LLMs) \cite{pan2023unifying}, leading to LLM-augmented medical KGs. However, this direction must be pursued cautiously to avoid undesirable issues, such as hallucinations or factual inaccuracies in the inferred relations.

Lastly, our current approach modeled the historical medication records by forming detached sub-graphs of clinical and medicine KGs for each admission. An appealing future direction would be to explore a unified medical KG that integrates clinical and medicine nodes, possibly bridged by a personalized patient node, and harnesses an even broader range of relations between medical entities with more robust relation-aware GNNs. 

\section{Conclusion}
\label{section:conclusion}

This study addressed the medicine recommendation task by proposing a novel graph-driven model called KindMed, which induces knowledge arising from multiple relations merged upon the EHR cohort. We combined observed medical codes and the extracted relations from external knowledge to construct personalized medical KGs and further enriched their node embeddings through relation-aware GNNs. We then devised a hierarchical sequence mechanism with a collaborative filtering layer that accounts for patients' medication histories by learning and fusing temporal dynamics from clinical and medicine streams across their longitudinal admissions. Our attentive prescribing module enhanced KindMed's predictive performance with an effective attention mechanism for associating the learned joint historical medical records, clinical progression, and patients' most recent clinical state representations. Finally, comprehensive ablation studies and experiments on KindMed variants using the MIMIC cohorts demonstrated improved performances compared to several closely related graph-driven medicine recommender baselines in the literature.

\section*{Acknowledgments}
This work was supported by National Research Foundation of Korea (NRF) grant funded by the Korea government (MSIT) No. 2022R1A2C2006865 (Development of deep learning techniques for data-driven medical knowledge graph generation and interpretable multi-modal electronic health records analysis). This work was further supported by Institute of Information \& communications Technology Planning \& Evaluation (IITP) grant funded by the Korea government (MSIT) No. 2019-0-00079 (Artificial Intelligence Graduate School Program (Korea University)) and No. 2022-0-00959 ((Part 2) Few-Shot Learning of Causal Inference in Vision and Language for Decision Making).

\bibliographystyle{IEEEtran}
\bibliography{main}


\begin{IEEEbiography}[{\includegraphics[width=1in,height=1.25in,clip,keepaspectratio]{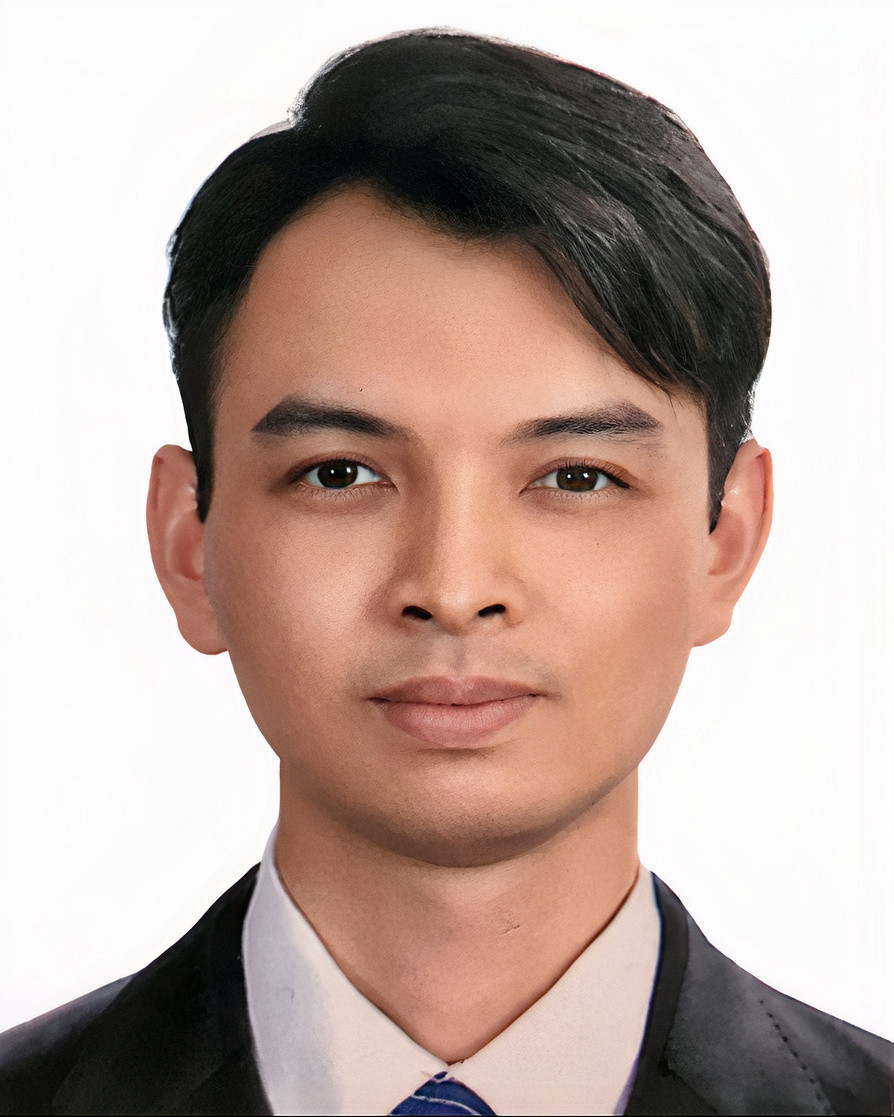}}]{Ahmad Wisnu Mulyadi}
received a bachelor’s degree in computer science education from the Indonesia University of Education, Bandung, Indonesia, in 2010.  

He is currently pursuing a Ph.D. degree with the Department of Brain and Cognitive Engineering, Korea University, Seoul, South Korea. His current research interests include machine/deep learning in healthcare, biomedical image analysis, and graph representation learning.
\end{IEEEbiography}

\begin{IEEEbiography}[{\includegraphics[width=1in,height=1.25in,clip,keepaspectratio]{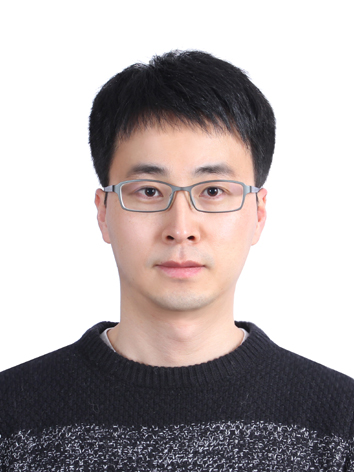}}]{Heung-Il Suk}(Senior Member, IEEE) is currently a Professor at the
Department of Artificial Intelligence and an Adjunct Professor at the Department of Brain and
Cognitive Engineering at Korea University. He
was a Visiting Professor at the Department of
Radiology at Duke University between 2022 and
2023.

He was awarded a Kakao Faculty Fellowship
from Kakao and a Young Researcher Award
from the Korean Society for Human Brain Mapping (KHBM) in 2018 and 2019, respectively. His
research interests include causal machine/deep learning, explainable
AI, biomedical data analysis, and brain-computer interface.

Dr. Suk serves as an Editorial Board Member for Clinical and Molecular Hepatology (Artificial Intelligence Sector), Electronics, Frontiers in
Neuroscience, Frontiers in Radiology (Artificial Intelligence in Radiology), International Journal of Imaging Systems and Technology (IJIST),
and a Program Committee or a Reviewer for NeurIPS, ICML, ICLR,
AAAI, IJCAI, CVPR, MICCAI, AISTATS, \etc.
\end{IEEEbiography}
\vfill

\end{document}